\documentclass{article} 
\usepackage[preprint]{colm2026_conference}
\usepackage{wrapfig}
\usepackage{microtype}
\usepackage{hyperref}
\usepackage{url}
\usepackage{booktabs}
\usepackage{subfig}
\usepackage{microtype}
\usepackage{graphicx}
\usepackage{booktabs} 
\usepackage{amsmath}
\usepackage{amssymb}
\usepackage{mathtools}
\usepackage{amsthm}
\usepackage{url}
\usepackage[utf8]{inputenc}
\usepackage[T1]{fontenc}
\usepackage{nicefrac}
\usepackage[table,xcdraw]{xcolor}
\usepackage{listings}
\usepackage{multirow}
\usepackage{colortbl}
\usepackage{xspace}
\usepackage{bbding}
\usepackage{wrapfig}
\usepackage{tcolorbox}
\tcbuselibrary{breakable}
\tcbuselibrary{skins}
\usepackage{tabularx}
\usepackage{makecell}
\usepackage{pifont}
\usepackage{float}
\usepackage{fancyvrb}
\usepackage{lipsum}
\usepackage{caption}
\usepackage{subcaption}
\usepackage{placeins}
\usepackage{enumitem}
\usepackage{epstopdf}

\usepackage{titletoc}
\usepackage{lineno}

\definecolor{darkblue}{rgb}{0, 0, 0.5}
\hypersetup{colorlinks=true, citecolor=darkblue, linkcolor=darkblue, urlcolor=darkblue}

\usepackage[capitalize,noabbrev]{cleveref}

\definecolor{myred}{RGB}{184,26,15}
\definecolor{mydarkred}{rgb}{0.6,0,0}
\definecolor{myblue}{HTML}{268BD2}
\newcommand{\gcmark}{\textcolor[RGB]{18,220,168}{\checkmark}}
\newcommand{\rxmark}{\textcolor[RGB]{202,12,22}{\ding{55}}}

\newtheorem*{Pro*}{Problem}

\newcommand{\High}{\textcolor[RGB]{247,85,115}{\gcmark~}}
\newcommand{\Medium}{\textcolor[RGB]{100,188,216}{$\triangle$~}}
\newcommand{\Low}{\textcolor[RGB]{117,196,119}{\rxmark~}}

\usepackage[textsize=tiny]{todonotes}

\def\method{\texttt{{Nano-Memory}}\xspace}

\title{Back to Basics: Let Conversational Agents Remember with Just Retrieval and Generation}

\makeatletter
\def\thanks#1{\protected@xdef\@thanks{\@thanks
        \protect\footnotetext{#1}}}
\makeatother

\author{Yuqian Wu\textsuperscript{1, *},\ \ Wei Chen\textsuperscript{1, *, $\dagger$},\ \ Zhengjun Huang\textsuperscript{2},\ \ Junle Chen\textsuperscript{2},\ \ Qingxiang Liu\textsuperscript{1},\\ 
{\bf Kai Wang\textsuperscript{3}, \ \ Xiaofang Zhou\textsuperscript{2},\ \ Yuxuan Liang\textsuperscript{1, $\dagger$}\thanks{\textsuperscript{*}Equal contribution. \textsuperscript{$\dagger$}Corresponding authors.}} \\ \\
The Hong Kong University of Science and Technology (Guangzhou)\textsuperscript{1} \\
The Hong Kong University of Science and Technology\textsuperscript{2} \\
National University of Singapore\textsuperscript{3} \\ \\
\textit{ywu188@connect.hkust-gz.edu.cn, \textsuperscript{$\dagger$}onedeanxxx@gmail.com, \textsuperscript{$\dagger$}yuxliang@outlook.com} \\
\textit{Code available at: \url{https://github.com/yuqian2003/Nano-Memory}}
\vspace{-4mm}
}

\begin{document}

\ifcolmsubmission
\linenumbers
\fi

\maketitle
\begin{abstract}
Existing conversational memory systems rely on complex hierarchical summarization or reinforcement learning to manage long-term dialogue history, yet remain vulnerable to context dilution as conversations grow. In this work, we offer a different perspective: the primary bottleneck may lie not in memory architecture, but in the \textit{Signal Sparsity Effect} within the latent knowledge manifold. 
Through controlled experiments, we identify two key phenomena: \textit{Decisive Evidence Sparsity}, where relevant signals become increasingly isolated with longer sessions, leading to sharp degradation in aggregation-based methods; and \textit{Dual-Level Redundancy}, where both inter-session interference and intra-session conversational filler introduce large amounts of non-informative content, hindering effective generation.
Motivated by these insights, we propose \method, a minimalist framework that brings conversational memory back to basics, relying solely on retrieval and generation via Turn Isolation Retrieval (TIR) and Query-Driven Pruning (QDP). TIR replaces global aggregation with a max-activation strategy to capture turn-level signals, while QDP removes redundant sessions and conversational filler to construct a compact, high-density evidence set.
Extensive experiments on multiple benchmarks demonstrate that \method achieves robust performance across diverse settings, consistently outperforming strong baselines while maintaining high efficiency in tokens and latency, establishing a new minimalist baseline for conversational memory.
\end{abstract}

\section{Introduction}

Large language models enable conversational agents capable of long-term user interaction. A key requirement for such agents is the ability to remember past interactions and use them to inform future responses. This capability, often referred to as conversational memory~\citep{hu2025memory}, has become a fundamental component of personalized dialogue systems.

\begin{figure*}[t]
    \centering
    \vspace{-6mm}
    \includegraphics[width=1.0\textwidth]{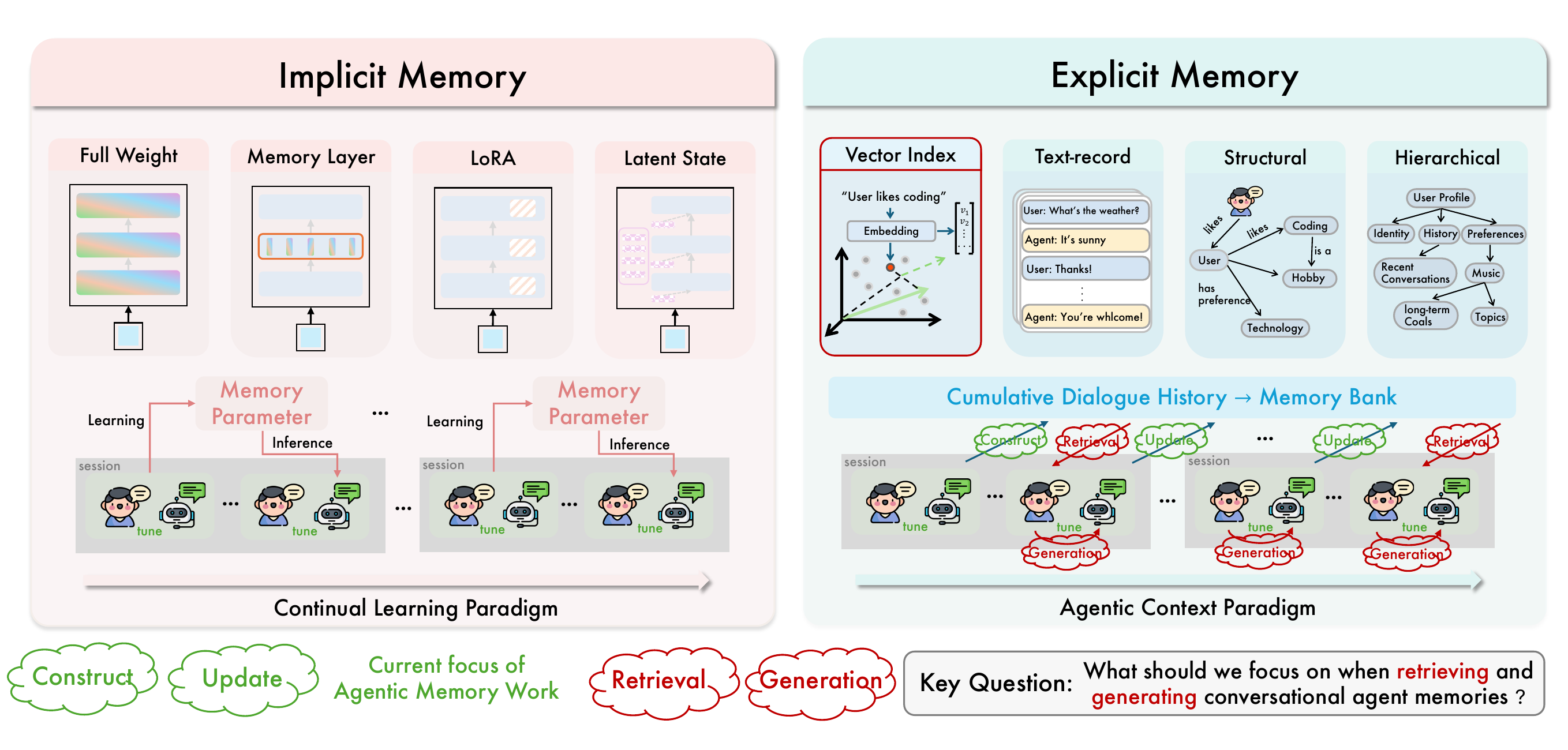}
    \vspace{-5mm}
    \caption{Different types of memory methods for personalized conversational agents}
    \label{fig.intro}
    \vspace{-6mm}
\end{figure*}

Existing research explores two primary paradigms for implementing conversational memory, as illustrated in Fig.~\ref{fig.intro}. One paradigm, implicit memory, integrates memory directly into the parameters or internal states of the language model. Representative approaches include updating full model weights~\citep{liu2024llm}, maintaining latent states~\citep{zhang2025memgen}, introducing dedicated memory layers~\citep{berges2024memory}, or relying on KV-cache~\citep{eyuboglu2025cartridges,zweiger2026fast} based persistence. These methods typically depend on continual learning mechanisms to encode user history into model parameters. Although promising, such approaches usually require additional training procedures, incur significant computational overhead, and may suffer from issues such as catastrophic forgetting or training instability~\citep{kemker2018measuring} inherent in neural networks.

As a result, another paradigm known as explicit memory remains the de facto solution in practice~\citep{claude2025memory,gpt2025memory}. In this setting, dialogue history is externalized into a memory bank that can be retrieved during inference. Recent studies (categorized in Appendix~\ref{appendix_baseline}, Tab.~\ref{tab:comparison}) have proposed increasingly sophisticated mechanisms for memory construction, organization, and updating, including vector indices, structured memories, hierarchical profiles, and text-based records. While these designs the usability of memory management, they also introduce considerable system complexity~\citep{anthropic2025harness,lopopolo2026harness}.


However, despite the growing sophistication of recent explicit memory designs, a fundamental question remains underexplored: \textit{what truly decisive evidence should both retrieval and generation focus on when serving as the core primitives for conversational memory?}

In this work, we provide a complementary perspective by examining how useful signals are distributed and utilized in conversational data. Our analysis reveals a \textit{Signal Sparsity Effect} in the latent knowledge manifold: only a small subset of turns contains evidence relevant to a query, while the majority of content is redundant. As conversations grow longer, this imbalance becomes increasingly severe, posing challenges to both retrieval and generation.


Through controlled experiments, we identify two key phenomena. First, \textit{Decisive Evidence Sparsity}: relevant signals become increasingly isolated as session length increases, making them difficult to capture with conventional aggregation-based retrieval. Second, \textit{Dual-Level Redundancy}: redundancy arises at both the inter-session level, where irrelevant sessions introduce interference, and the intra-session level, where conversational filler dominates most turns and dilutes useful information. Together, these significantly effects degrade both retrieval precision and generation quality. These findings highlight a simple yet overlooked principle: when using retrieval and generation as the foundation of conversational memory, the key is to \textit{accurately localize decisive signals and eliminate redundant context}.

Based on this insight, we back to the basics and propose \method, a minimalist framework built solely upon retrieval and generation. Specifically, we introduce \textit{Turn Isolation Retrieval} (TIR), which replaces coarse aggregation with a max-activation strategy to capture turn-level signals, and \textit{Query-Driven Pruning} (QDP), which removes redundant sessions and intra-session conversational filler to construct a compact, high-density evidence set for generation. Our contributions are summarized as follows:


\begin{itemize}[topsep=2pt, partopsep=0pt, leftmargin=6mm,parsep=2pt]
\item We identify the \textit{Signal Sparsity Effect} as a fundamental bottleneck in conversational memory and characterize it through two key phenomena: Decisive Evidence Sparsity and Dual-Level Redundancy.
\item We propose \method, a minimalist framework that returns conversational memory to the basic paradigm of retrieval and generation, instantiated via TIR and QDP.
\item We demonstrate through extensive experiments that a simple, well-designed retrieval-generation pipeline can outperform more complex memory systems, establishing a new minimalist baseline for long-term conversational memory.
\end{itemize}
\section{Preliminary}

\textbf{Definition 1.} \textit{(Conversational Interaction)}.
A conversational agent interacts with a user through a sequence of dialogue sessions: $\mathcal{H} = \{c_i\}_{i=1}^{C}$, where $C$ denotes the number of sessions in the dialogue history. Each session $c_i$ is composed of a sequence of interaction turns: $c_i = \{t_j\}_{j=1}^{T_i}$, where $T_i$ is the number of turns in session $i$. Each turn is defined as a pair: $t_j = (u_j, r_j)$, where $u_j$ denotes the user request $r_j$ denotes the agent response. The complete interaction history $\mathcal{H}$ therefore contains all previously observed user–agent interactions accumulated across sessions.

\textbf{Definition 2.} \textit{(Explicit Conversational Memory)}.
Given a dialogue history $\mathcal{H}$, an explicit conversational memory is an external memory bank constructed from historical interactions without modifying the parameters of the response model. Formally, the memory bank is defined as $\mathcal{M}=\{m_k\}_{k=1}^{|\mathcal{M}|}$, where each memory unit $m_k$ represents a textual fragment derived from the dialogue history $\mathcal{H}$. The initial memory bank is produced by a \textit{memory construction function} $f_c:\mathcal{H}\rightarrow\mathcal{M}$, which transforms the dialogue history into a collection of retrievable memory entries. Furthermore, as the conversation progresses dynamically, this memory bank requires continuous maintenance governed by a \textit{memory update function} $f_u:(\mathcal{M}, \Delta h)\rightarrow\mathcal{M}'$, where $\Delta h$ denotes newly arriving dialogue turns or newly extracted user insights. The design of $f_c(\cdot)$ and $f_u(\cdot)$ collectively determines how conversational history is partitioned, organized, indexed, and maintained over time.

Existing studies primarily focus on complex construction and update functions, employing structured organizations like hierarchical~\citep{hu2025hiagent,rezazadeh2025isolated} or graph-based memories~\citep{zhang2025gmemory}, alongside update strategies such as summarization~\citep{xu2025amem, tan2025prospect, yao2026arc} or dynamic forgetting~\citep{packer2024memgpt, zhang2025memevolve, yue2026mem}. However, maintaining these intricate structures introduces significant computational overhead and risks losing fine-grained contextual nuances. Diverging from these heavily engineered paradigms, we simply append raw historical interactions, allowing the memory bank $\mathcal{M}$ to grow naturally in an unstructured form. By preserving unaltered conversational contexts, we pivot to a query-centric perspective, focusing entirely on the challenges of the retrieval and augmentation stages.

\textbf{Problem.} \textit{(Explicit Memory-Augmented Conversational Agent)}.
Given a user request $u^*$, a budget of $N$ context units, and a memory bank $\mathcal{M}$, the goal is to generate an accurate response $r^*$ by leveraging the external memory bank in two stages: \textit{(\romannumeral1) Retrieval}, where a retrieval function $f_r$ identifies the $N$ most relevant memory units $\{m_n\}_{n=1}^N \leftarrow f_r(u^*, \mathcal{M}, N)$ based on the query $u^*$; and \textit{(\romannumeral2) Generation}, where the retrieved units are organized chronologically and fed into a language agent to synthesize the final response $r^* = f_{g}(u^*, \{m_n\}_{n=1}^N)$.

This formulation aligns with the general paradigm of retrieval-augmented generation~\citep{lewis2020retrieval}, where an external memory source provides additional grounding information for the language agent’s responses, but differs in that the retrieval corpus consists of the agent’s accumulated historical interactions rather than a static external knowledge base.

\section{Methodology}~\label{sec:method}
\vspace{-8mm}
\subsection{Associate Deep with Turn Isolation Retrieval}

\textbf{Intuition.} As mentioned above, we argue that the bottleneck of long-term memory lies not in organization, but in the activation mechanism within the latent knowledge manifold. As the conversation history $\mathcal{H}$ expands, it forms an increasingly dense manifold of interactions. We hypothesize that within this manifold, the information critical to a specific query is not distributed uniformly or densely, but rather exhibits pronounced \textit{decisive evidence sparsity}.

\begin{figure*}[t!]
  \centering
    \includegraphics[width=1.0\textwidth]{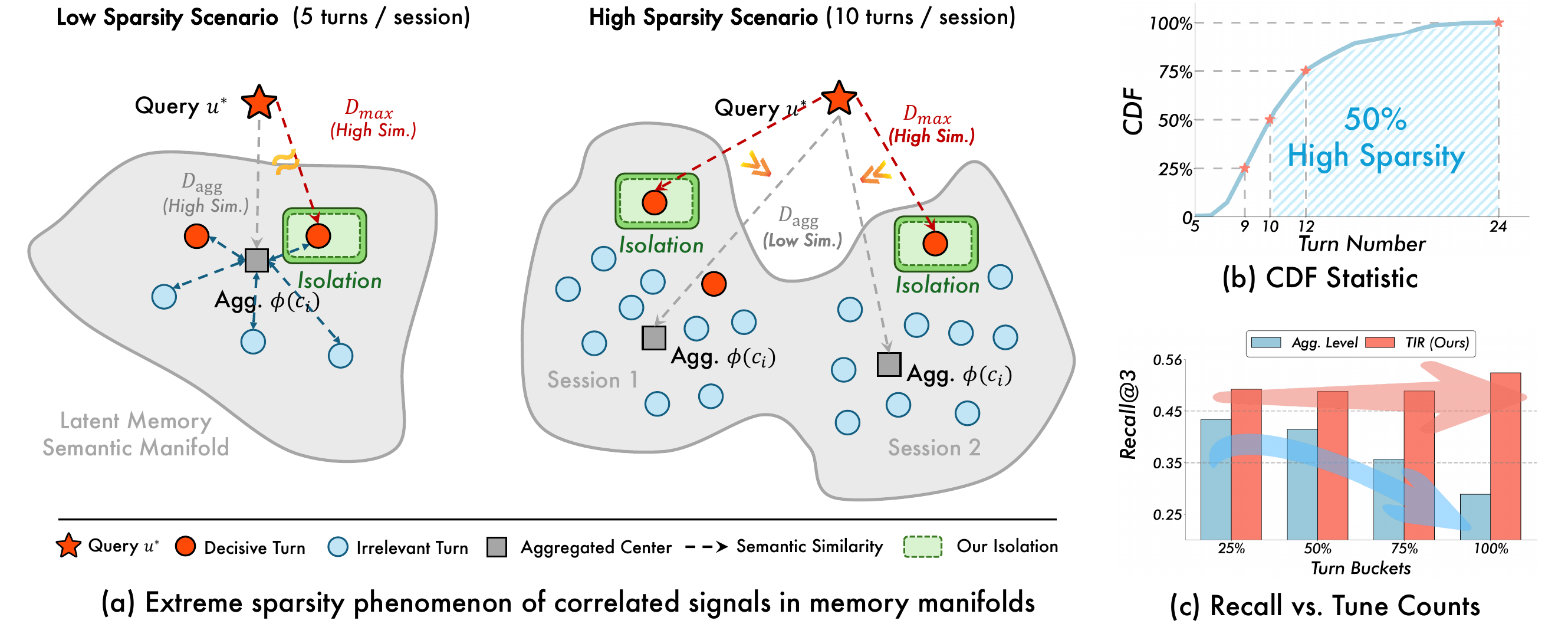}
  \vspace{-2mm}  
  \caption{Illustration of \textit{Signal Sparsity Effect} in latent memory manifolds and the robust retrieval performance of \textit{TIR} across varying session lengths.}
  \label{fig:method_1} 
  \vspace{-4mm}
\end{figure*}

\textbf{Empirical Study.}
To validate this hypothesis, we conduct a statistical analysis on the LoCoMo dataset, focusing on the semantic relevance between queries $u^*$ and individual turns $t_{i,j}$ across varying session scales. We observed the following findings from the experiment:

\textit{\ding{88} Finding \uppercase\expandafter{\romannumeral1}: The Sparsity of Decisive Evidence.}
We first examined overall session length, as shown in Fig.\ref{fig:method_1}~(b), which illustrates the cumulative distribution function of tune number. While a small percentage of sessions contain fewer than 5 tunes, a significant proportion (up to 50\%) have 10 or more tunes. As shown in Fig.\ref{fig:method_1}~(a), in compact sessions, relevant signals are relatively concentrated. However, in extended sessions, the decisive turns (high-relevance points) become geographically isolated within a sea of irrelevant contexts.

\textit{\ding{88} Finding \uppercase\expandafter{\romannumeral2}: The Failure of Global Aggregation.} A direct consequence of sparsity is context dilution. Traditional paradigms, based on aggregation~\citep{memgas,secom}, attempts to represent session $c_i$ by compressing the entire content of the session into a vector $\phi(c_i)$, as shown in Fig.\ref{fig:method_1}~(a). In compact sessions, this remains close to decisive evidence. However, in longer sessions, this naturally diverges and becomes trivial. Furthermore, as shown in Fig.\ref{fig:method_1}~(c), while aggregation-based "turning point level" retrieval performs well in short sessions (25th percentile), its Recall@3 drops sharply with increasing session length (from approximately 0.44 to <0.30). This confirms that the global mean deviates significantly from the target due to the increase in irrelevant words, leading to a failure in retrieval scope.

\textbf{Implementation Details.}
Motivated by the verified sparsity, we move away from "over-engineered" memory structures and propose the \textit{Turn Isolation Retrieval (TIR)} mechanism. Instead of attempting to suppress noise through complex summarization, TIR bypasses dilution by directly isolating high-evidence signals within the manifold. 

Specificity, Given a memory bank $\mathcal{M}$ and a query $u^*$, we employ a embedding model $\phi$ to project inputs into the latent semantic space. For each constituent turn $t_{i,j}$ within a candidate session $c_i$, we calculate its fine-grained local relevance: 
$s(u^*, t_{i,j}) = \langle \phi(u^*), \phi(t_{i,j}) \rangle$.
To prevent the decisive signal from being obscured by the intra-session context, we define the overall session relevance $\mathcal{S}$ through a max-activation strategy:
$\mathcal{S}(c_i, u^*) = \max_{1 \le j \le T_i} s(u^*, t_{i,j})$.
Unlike mean-pooling, which is sensitive to session length $T_i$, this max-isolation ensures that the retrieval score is anchored solely to the most relevant evidence, maintaining a robust retrieval margin even in extremely long sessions.
Finally, the retrieval system $f_r$ ranks all candidate sessions and selects the Top-$k$ units:$\{ m_n \}_{n=1}^k = \operatorname*{Top-}k_{c_i \in \mathcal{M}} \mathcal{S}(c_i, u^*)$.

As shown in Fig.~\ref{fig:method_1}(c), our TIR mechanism exhibits remarkable stability across all session length percentiles, effectively neutralizing the impact of context expansion.

\vspace{2mm}
\subsection{Reply Sharp with Query Driven Pruning}
\vspace{1mm}

\textbf{Intuition.}
During retrieval, the TIR mechanism identifies high-gain signal points (Top-$k$ units) in the implicit memory manifold; however, following default settings, the retrieved content is typically the Top-$k$ conversation sessions corresponding to the Top-$k$ units. Therefore, our key intuition is that retrieval solves the "accessibility" problem in the generation phase, but not entirely the "understanding" problem: directly inputting these raw fragments hinders the efficient integration of sparse signals distributed across time slices into a globally coherent response. Furthermore, each dialogue unit ($c_i$) still contains a significant amount of discourse markers, verbal redundancy, and background information irrelevant to the current query ($u^*$). 
Injecting such content into the lengthy context generation process exacerbates noise and further degrades performance, constituting \textit{dual-level redundancy}.

\begin{figure*}[t!]
    \vspace{3mm}
  \centering
    \includegraphics[width=1.0\textwidth]{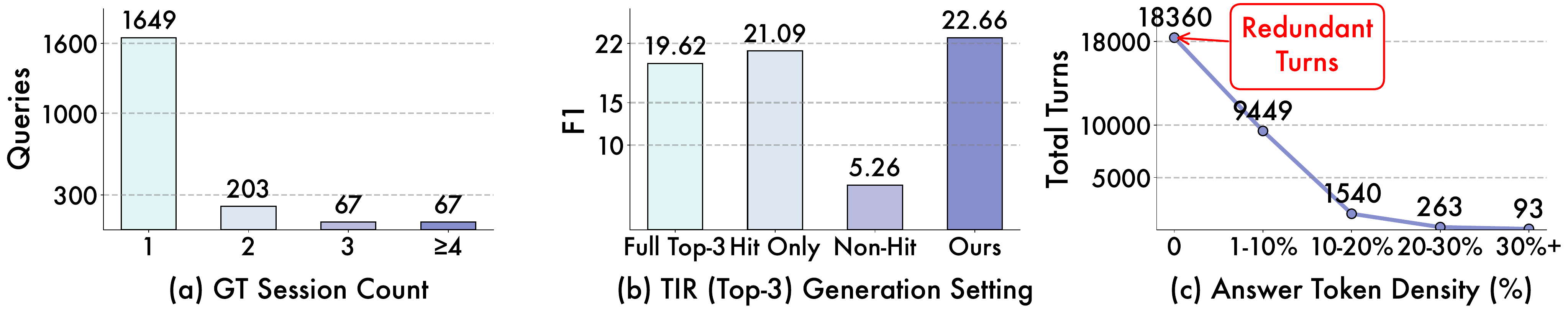}

  \vspace{2mm}  
  \caption{
    (a) Distribution of the Ground Truth (GT) session count required per query. (b) Downstream QA performance (F1) across different Top-$3$ generation settings. (c) Distribution of F1 support scores for individual turns within the retrieved GT sessions.
  }
  \label{fig:method_2} 
  \vspace{-8mm}
\end{figure*}

\textbf{Empirical Study.} 
To validate context refinement, we analyze how retrieved sessions impact response quality to guide our generation strategy, identifying two distinct sources of noise.

\textit{\ding{88} Finding \uppercase\expandafter{\romannumeral3}: The Inter-Session Interference Effect.}
We first examine the noise across sessions. As shown in Fig.~\ref{fig:method_2}(a), the vast majority of user queries inherently require evidence from only a single Ground Truth (GT) session. Consequently, a standard top-$k$ retrieval setup ($k>1$) inevitably introduces irrelevant sessions into the generation stage.
Fig.~\ref{fig:method_2}(b) explicitly demonstrates this negative impact: generating answers using only the correctly retrieved GT session ("Hit Only") achieves an F1 score of 21.09, which significantly outperforms the score obtained using the complete retrieved sessions ("Full Top-3").
Furthermore, forcing the LLM to generate using only the irrelevant Top-3 sessions ("Non-Hit") decreases the F1 score to 5.26, demonstrating that such redundant context acts as misleading semantic noise.

\textit{\ding{88} Finding \uppercase\expandafter{\romannumeral4}: Intra-Session Conversational Redundancy.}
Even when $f_{r}$ successfully isolates GT sessions, significant redundancy still exists.
To illustrate this, we aggregate all constituent turns from ground truth sessions in LoCoMo (totaling 29,705 turns) and evaluate their individual information density. 
As shown in Fig.~\ref{fig:method_2}(c), we measure the \textit{Answer Token Density} by computing the token-level F1 percentage between each raw turn and the ground-truth answer to quantify the exact word-match evidence.The distribution reveals that out of all turns within the ground truth sessions, a massive 18,360 turns yield zero overlap.
Yielding no semantic contribution, these turns act purely as redundant conversational padding (e.g., greetings, topic transitions) that misguides generation.
Furthermore, even the minority of turns that contain answer tokens predominantly fall into the lowest density bin (1\%-10\%), as the critical information is heavily diluted by conversational filler. This heavily skewed distribution intuitively demonstrates that ground truth sessions remain overwhelmingly dominated by off-topic conversational inertia rather than query-relevant facts.

\textbf{Implementation Details.}
Motivated by these findings, we identify intra- and inter-session redundancy as the primary semantic bottlenecks for generation. To resolve this, we propose \textit{Query Driven Pruning (QDP)} mechanism.
Following the retrieval top-($k$) sessions $\mathcal{C}_{ret} = \{c_1, \dots, c_k\}$, we consolidate them into a unified context $\mathcal{E}_{fused}$, then employ a filtering language model $\mathcal{M}_{f}$ to process. 
Specifically, $\mathcal{M}_{f}$ is prompted (see Appendix~\ref{appendix_prompts}) to extract only the query-relevant fragments from $\mathcal{E}_{fused}$.This operation yields the distilled context $\hat{\mathcal{H}} = \text{Filter}(\mathcal{M}_{f}, \mathcal{E}_{fused}, u^*)$, explicitly purging both inter- and intra-session noise.
Finally, the response $a$ is synthesized by the generator $f_g$ using only the pruned context: $a = f_g(\tilde{\mathcal{H}}, u^{*})$.


As shown in Fig~\ref{fig:method_2}(b), QDP achieves an F1 score of 22.66, significantly outperforming the "Hit Only" setting. This confirms that beyond merely isolating the ground truth session, QDP effectively mitigates intra-session redundancy and maximizes memory utility.

\section{Experiments}
In this section, we conduct extensive experiments to study the following research questions:

\begin{itemize}[topsep=1pt, partopsep=0pt, leftmargin=2mm,parsep=1pt]
\item \textbf{RQ1: (Effectiveness)} Can \method outperform existing agent memory methods?

\item \textbf{RQ2: (Universality)} Can \method maintain consistent improvements across different language backbones, retrieval models and diverse query types?

\item \textbf{RQ3: (Efficiency)} How does \method compare to baselines in terms of computational overhead, including construction time (offline), inference latency (online), and overall token consumption?

\item \textbf{RQ4: (Ablation \& Mechanism)} How does each component of \method contribute to its overall performance?
\end{itemize}

\textbf{Benchmarks}
We conduct experiments on four publicly available benchmarks: LoCoMo~\citep{locomo}, Long-MT-Bench+~\citep{secom}, LongMemEval-s~\citep{longmemeval} and LongMemEval-m~\citep{longmemeval}, which are specifically designed to evaluate agent capabilities in managing long-term conversational dependencies.
We utilize the complete set of QA pairs for evaluation, with detailed statistics listed in Appendix~\ref{appendix_datasets}. 

\textbf{Baselines} 
We compare \method with Full History and representative memory baselines:
MPNet~\citep{song2020mpnet}, Contriever~\citep{contriever},MPC~\citep{MPC},RecurSum~\citep{RecurSum},SeCom~\citep{secom},HippoRAG 2~\citep{hipporag2},RAPTOR~\citep{raptor},A-Mem~\citep{xu2025amem} and MemGAS~\citep{memgas}.
More baseline details in Appendix~\ref{appendix_baseline}

\textbf{Protocol.} We evaluate model performance using F1~\citep{locomo}, BLEU~\citep{papineni2002bleu}, ROUGE~\citep{chin2004rouge},  BERTScore~\citep{zhang2020bertscore} and GPT-4o-as-Judge (4o-J)~\citep{zheng2023judging}. 
Following the standard experimental setup~\citep{memgas},we use \texttt{gpt-4o-mini-2024-07-18} (temperature=0) as generator and Contriever as retriever. 
All baselines operate with a consistent top-$k$ ($k=3$) retrieval and unified prompts (see Appendix~\ref{appendix_prompts}). We exclude LongMTBench+ due to missing retrieval ground truth and the retrieval performance of RAPTOR and A-Mem cannot be evaluated in all tasks. Results for RAPTOR, A-Mem, and HippoRAG v2 on LongMemEval-m are omitted due to high computational costs. The best and second-best results are in \textbf{bold} and \underline{underlined}, respectively.

\subsection{Effectiveness Analysis}

\begin{table*}[htbp!]
\setlength\tabcolsep{0pt}  
\renewcommand{\arraystretch}{0.9} 
\small
\centering
\caption{Retrieval Performance on LoCoMo. (Full results can be found in Appendix Table~\ref{tab:appendix_retrieval_whole}.)}
\vspace{-2mm}
\label{tab:retrieval}
\begin{tabular*}{\textwidth}{@{\extracolsep{\fill}}@{} l|cccccc|c @{}}
\toprule
\textbf{Model} &\textbf{Recall@3 }&\textbf{NDCG@3}&\textbf{Recall@5} &\textbf{NDCG@5}&\textbf{Recall@10} & \textbf{NDCG@10} & \textbf{Time}\\

\midrule
MPNet ~\citeyearpar{song2020mpnet}& 45.92 & 47.68 & 53.98 & 51.76 & 68.58 & 56.85 & 1.42s\\

Contriever ~\citeyearpar{contriever} &49.95& 52.15 &58.31& 56.29& 71.80& 60.91 & 1.24s\\

MPC ~\citeyearpar{MPC} & 49.50  & 51.47  & 57.45  & 55.53  & 71.85  & 60.47 & 0.81s \\

RecurSum ~\citeyearpar{RecurSum} &47.23&48.99&59.01&54.58&74.97&60.07 & 1.26s\\

SeCom ~\citeyearpar{secom} & {52.97}& {54.03}& {64.05}& {59.37}& {78.25}& {64.28}& 2.17s\\

HippoRAG 2~\citeyearpar{hipporag2} &56.60&58.37&65.06&62.50&78.05&66.79& 3.70s\\

MemGAS~~\citeyearpar{memgas}  &\underline{56.85}&  \underline{59.22}&  \underline{67.98}& \underline{64.25}&   \underline{81.22}& \underline{68.71}  & 2.64s \\

\midrule

\shortstack{\method \\ \vspace{0.3mm}} & 
\shortstack{\textbf{69.39} \\ \vspace{-1mm} \\ {\tiny \textcolor[HTML]{878ECD}{($\uparrow$22.06\%)}}} & 
\shortstack{\textbf{71.19} \\ \vspace{-1mm} \\ {\tiny \textcolor[HTML]{878ECD}{($\uparrow$20.21\%)}}} & 
\shortstack{\textbf{76.33} \\ \vspace{-1mm} \\ {\tiny \textcolor[HTML]{878ECD}{($\uparrow$12.28\%)}}} & 
\shortstack{\textbf{74.58} \\ \vspace{-1mm} \\ {\tiny \textcolor[HTML]{878ECD}{($\uparrow$16.08\%)}}} & 
\shortstack{\textbf{86.46} \\ \vspace{-1mm} \\ {\tiny \textcolor[HTML]{878ECD}{($\uparrow$6.45\%)}}} & 
\shortstack{\textbf{77.99} \\ \vspace{-1mm} \\ {\tiny \textcolor[HTML]{878ECD}{($\uparrow$13.51\%)}}} & 
\shortstack{\textbf{0.74s} \\ \vspace{-1mm} \\ {\tiny \textcolor[HTML]{878ECD}{($\downarrow$71.97\%)}}} \\

\bottomrule

\end{tabular*}
\end{table*}

To evaluate the overall effectiveness of our proposed method, we conducted a comprehensive comparison with various ten baselines on four long-term conversation benchmarks. The performance of QA generation and retrieval are presented in Tab.~\ref{tab:retrieval} and Tab.~\ref{tab:QA performance}, respectively. 
We observe \ding{182} \textit{Outstanding QA Generation}: 
\method achieves competitive generation performance compared to existing baselines. For example, on the LoCoMo dataset, \method~ delivers a significant 28.3\% improvement in F1 score over the strongest baseline. 
\ding{183} \textit{Competitive Retrieval}:  
Tab.~\ref{tab:retrieval} indicates the competitive performance advantages of \method in the retrieval stage, outperforming the strongest baseline by 22.05\% in Recall$@$3 on LoCoMo dataset. Notably, on LongMemEval-s dataset (detailed in Appendix Tab.~\ref{tab:appendix_retrieval_whole}), although MemGAS achieves higher retrieval recall, its multi-granularity mechanism inherently introduces broad conversational noise during the retrieval stage, which ultimately degrades its final QA performance. In contrast, \method utilizes QDP during the generation stage to filter out intre- and intra-session noise, effectively leveraging raw retrieval results to achieve superior downstream QA performance.

\begin{table*}[t!]
\setlength\tabcolsep{0pt}
\small
\renewcommand{\arraystretch}{0.8}
\centering
\caption{QA performance. Contriever is the retrieval backbone (excluding Full History and MPNet), with \texttt{gpt-4o-mini-2024-07-18} as the generator. \textit{Avg.Tokens} denotes the average token consumption during the generation phase.} 
\vspace{-2mm}

\begin{tabular*}{\linewidth}{@{\extracolsep{\fill}} l|ccccccc |cc }

\toprule
\textbf{Model} &\textbf{4o-J}  &\textbf{F1} & \textbf{BLEU4}& \textbf{ROUGE1}&\textbf{ROUGE2}&\textbf{ROUGEL} & \textbf{BERTScore} & \textbf{\makecell{Avg.\\Tokens}}\\ 
\midrule
\multicolumn{10}{c}{\cellcolor{cyan!5}\textit{\textbf{LoCoMo}}} \\
\midrule
Full History &33.43&12.23&1.84&12.70&5.66&11.73&84.07 & 20,078\\
MPNet ~\citeyearpar{song2020mpnet} & 38.07 & 14.52 & 2.36 & 14.97 & 6.82 & 13.84 & 84.46 & 2,474\\

Contriever ~\citeyearpar{contriever}&40.33&15.76&2.77&16.08&7.75&15.10&84.70 & 2,348\\

MPC ~\citeyearpar{MPC} & 40.38 & 14.81 & 1.99 & 15.10 &6.83 & 14.13 & 84.43 & 2,683\\

RAPTOR ~\citeyearpar{raptor} & 29.46 &14.44&2.85&14.97&7.37&14.02& 84.39 & 1,931\\

RecurSum ~\citeyearpar{RecurSum} & 22.56&9.14&0.99&9.82&3.38&8.98&83.45 & 3,074\\

HippoRAG 2~\citeyearpar{hipporag2} & \underline{45.62}& 14.95 & 2.76 & 15.38 & 7.52 & 14.35 & 84.50 & 2,991 &\\
A-Mem ~\citeyearpar{xu2025amem} &40.81 & 14.72 & 2.83 & 16.22 & 7.71 & 14.89&84.72 & 3,042\\
SeCom ~\citeyearpar{secom} & 43.45&15.28& 3.12 & 17.16 & 8.52 & 16.01 & 84.84 & 1,021\\ 
MemGAS~\citeyearpar{memgas}  &41.07&\underline{17.66}&\underline{3.61}&\underline{18.00}&\underline{8.93}&\underline{16.99}&\underline{85.13} & 2,825\\

\method & \textbf{48.84} & \textbf{22.66} & \textbf{5.33} & \textbf{22.92} & \textbf{12.04} & \textbf{21.68} & \textbf{85.94} & 1,403\\

\midrule
\multicolumn{10}{c}{\cellcolor{cyan!5}\textit{\textbf{LongMTBench+}}} \\
\midrule
Full History & \underline{67.44}&36.07&11.32&37.90&20.51&29.25&87.81 & 19,194\\
MPNet ~\citeyearpar{song2020mpnet}&63.89&35.32&11.10&37.32&20.45&28.57&87.73 & 12,187\\

Contriever ~\citeyearpar{contriever}&63.54& 31.82 &7.51&33.42&18.91 & 26.53 & 87.01 & 12,045 \\
MPC ~\citeyearpar{MPC} &61.81&31.52&7.97&33.56&17.27&25.20&86.51 & 12,289\\
RAPTOR ~\citeyearpar{raptor} & 59.72 & 37.69 & 13.47 & 40.08 & 21.68 & 30.88 & 88.38 & 10,631\\
RecurSum ~\citeyearpar{RecurSum} & 24.65&26.58&6.91&29.23&11.93&20.90&86.11 & 13,527\\
HippoRAG 2~\citeyearpar{hipporag2} & 63.54&35.64&11.05&37.61&20.37&28.76&87.70 & 13,583\\
A-Mem ~\citeyearpar{xu2025amem} & \underline{65.73} & 36.82 & 11.36& 38.92 & 20.88 & 29.14 & 87.92 & 13,735\\
SeCom ~\citeyearpar{secom} & 64.58 & 38.89 & 13.40 & 40.87 & 22.81 & 31.90 & 88.45 & 4,714\\ 

MemGAS~\citeyearpar{memgas} & \textbf{67.71} & \underline{41.49} & \underline{15.78} & \underline{43.69} & \underline{24.47}& \underline{34.64}& \underline{88.90}& 12,873\\

\method & 64.15 & \textbf{42.40} & \textbf{16.83} & \textbf{44.55} & \textbf{25.45} & \textbf{35.57} & \textbf{89.34} & 6,174 \\ 

\midrule
\multicolumn{10}{c}{\cellcolor{cyan!5}\textit{\textbf{LongMemEval-s}}} \\
\midrule
Full History & 50.60&11.48&1.40&12.10&5.47&10.85&83.07 & 103,137  \\

MPNet ~\citeyearpar{song2020mpnet}&53.20&13.96&2.21&14.49&6.78&12.93&83.72 & 8,173  \\

Contriever ~\citeyearpar{contriever}& 55.40 & 13.37 & 2.15 & 13.90 & 6.72 & 12.43 & 83.64 & 8,471 \\

MPC ~\citeyearpar{MPC} &53.80&13.60&1.74&14.27&6.49&12.95&83.49 & 8,457 \\
RAPTOR ~\citeyearpar{raptor} & 32.20&12.08&1.90&12.73&5.82&11.25&83.50& 6,254  \\
RecurSum ~\citeyearpar{RecurSum} & 35.40&12.29&2.09&13.01&5.55&11.52&83.60 & 8,853  \\
HippoRAG 2~\citeyearpar{hipporag2} & 57.60 & 14.73 &2.15 & 15.30 & 7.36 & 13.83 & 83.86 & 8,530 \\
A-Mem \citeyearpar{xu2025amem} & 55.60 & 13.73 & 2.11 & 14.82 & 6.81 & 12.98 & 83.88 &9,018  \\
SeCom ~\citeyearpar{secom} & 56.00 & 12.95& 2.25 &13.80&6.09&11.93&83.51 &2,741  \\

MemGAS ~\citeyearpar{memgas}&\textbf{60.20}&\underline{20.74}&\underline{4.48}&\underline{21.36}&\underline{10.47}&\underline{19.80}&\underline{85.29} &8,829 \\

\method & \underline{57.20} & \textbf{21.06} & \textbf{4.89} & \textbf{21.80} & \textbf{10.77} & \textbf{20.16} & \textbf{85.54} & 4,302\\

\midrule
\multicolumn{10}{c}{\cellcolor{cyan!5}\textit{\textbf{LongMemEval-m}}} \\
\midrule
Full History & 12.20&5.70&0.78&6.27&2.08&5.28&81.62 & 128,000\\

MPNet ~\citeyearpar{song2020mpnet}&37.80&10.97&1.65&11.57&4.99&10.07&83.06& 7,989\\

Contriever ~\citeyearpar{contriever}&42.80&11.88&1.66&12.63&5.51&11.11&83.31 & 8,274\\

MPC ~\citeyearpar{MPC} &37.80&11.28&1.37&11.93&5.12&10.57&82.98&8,428\\

RecurSum ~\citeyearpar{RecurSum} &23.80&10.04&1.70&10.89&4.26&9.21&83.12 & 8,927\\

SeCom ~\citeyearpar{secom} & 42.80&11.33&1.79&12.03&5.07&10.49&83.36 & 2,821\\

MemGAS~\citeyearpar{memgas} &\underline{45.40}&\underline{16.85}&\underline{3.39}&\underline{17.60}&\underline{8.25}&\underline{16.14}&\underline{84.69} & 8,852\\

\method & \textbf{46.60} & \textbf{18.09} & \textbf{3.96} & \textbf{18.88} & \textbf{9.03} & \textbf{17.40} & \textbf{85.10} & 4,261 \\ 

\bottomrule
\end{tabular*}
\label{tab:QA performance}
\end{table*}

\begin{figure*}[t!] 
    \centering
    \includegraphics[width=1.0\textwidth]{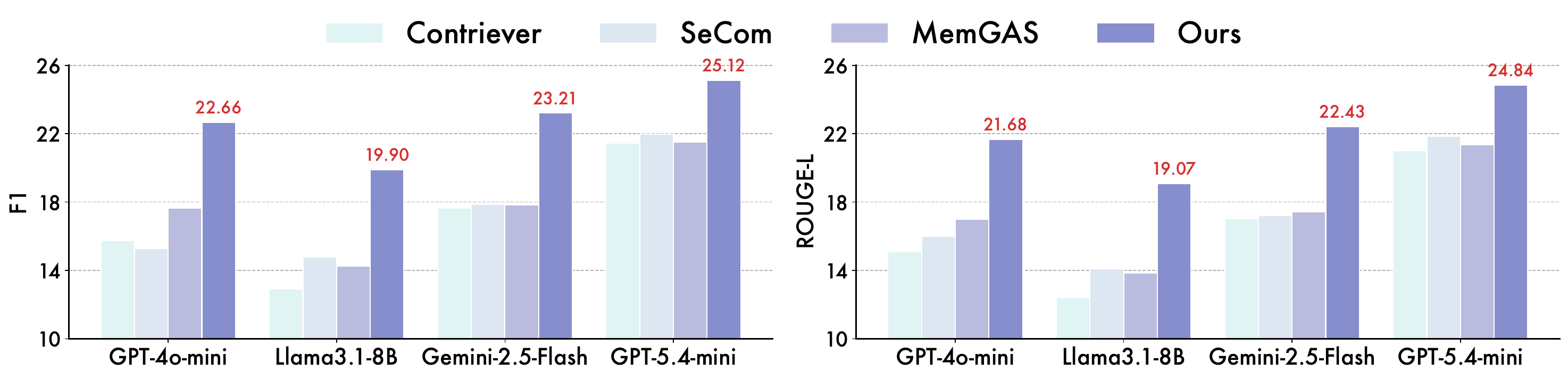}
    \caption{Comparison of different LLM backbones on LoCoMo. (Full results in Tab.~\ref{tab:diff_gen}.)}
    \label{fig:uni_llm}
    
    
    \includegraphics[width=1.0\textwidth]{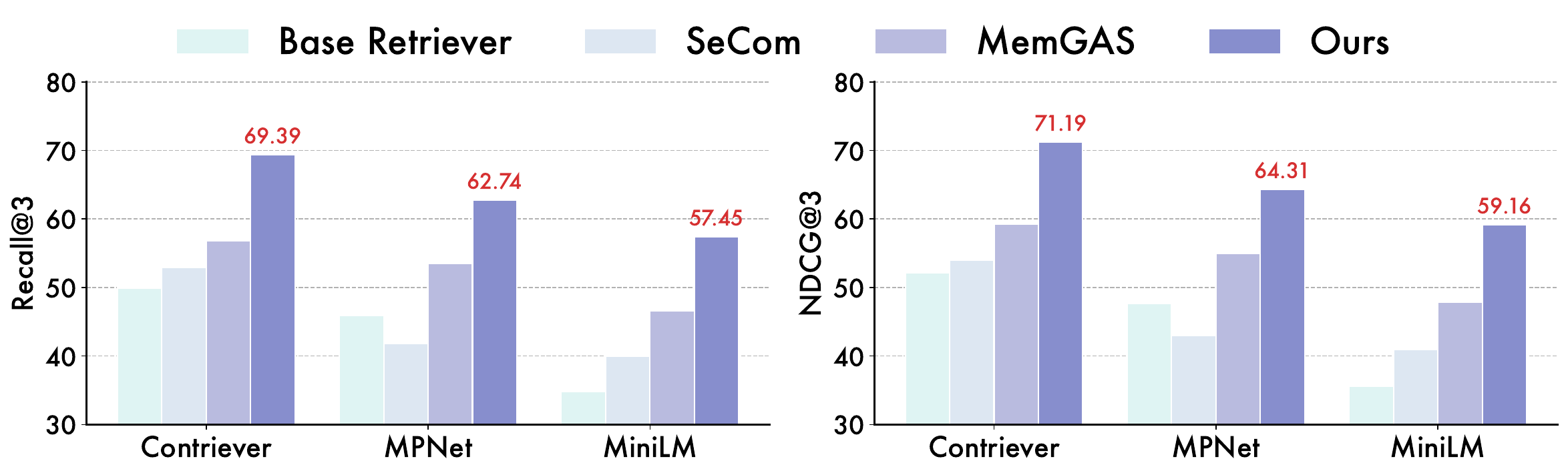}
    \caption{Comparison of different retrievers on LoCoMo. \textit{Base Retriever} refers to the vanilla Contriever, MPNet, and MiniLM. (Full results can be found in Appendix Tab.~\ref{tab:diff_retriever}.)}
    \label{fig:uni_retriever}
\end{figure*}

\subsection{Universality Analysis}

To comprehensively evaluate the robustness of \method, we conduct a universality analysis across three key dimensions:
\ding{182}\textit{Varying LLM backbones:} As shown in Fig.~\ref{fig:uni_llm}, \method demonstrates strong universality by achieving substantial performance gains with both closed-source and open-source LLMs.
\ding{183} \textit{Different retrievers:}Fig.~\ref{fig:uni_retriever} demonstrates that \method consistently achieves outstanding retrieval performance across various retrieval backbones. Specifically, it yields relative Recall$@$3 improvements of 22.06\%, 17.34\% and 23.20\% over the strongest baseline when paired with Contriever, MPNet, and MiniLM, respectively.
The corresponding final QA generation performance for each retriever is detailed in Appendix~\ref{appendix_more_exp_effectiveness} (Tab.~\ref{tab:diff_gen_retriever}).
\ding{184} \textit{Diverse query types} Fig.~\ref{fig:query} highlights \method achieves the best performance across all query types. 
Notably, the largest relative gain (39.7\%) is observed on \textit{Temporal} queries, directly exposing the inherent flaw of existing retrieval mechanisms that lose hard temporal constraints during context compression.
Results for different retrievers under various query settings are detailed in Appendix~\ref{appendix_more_exp_university} (Fig.~\ref{fig:whole_query_retriever}).

\subsection{Efficiency Analysis}

\begin{table*}[t]
\centering
\caption{Time Cost on LoCoMo under the same setting.}
\vspace{-2mm}

\begin{tabular*}{\linewidth}{@{\extracolsep{\fill}} l|c c c|c c}
\toprule
\textbf{Model} & {\small \textbf{Consturct}} & {\small \textbf{Retrieval}} & {\small \textbf{Generation}} & {\small \textbf{Total Time}} & \textbf{F1}\\
\midrule
Contriever & \textbackslash & 1.24s & 381.71s. & 382.95s & 15.76 \\
MPNet & \textbackslash & 1.42s & 317.28s & 318.70s & 14.52 \\
Secom & 294.62s & 2.17s & 111.81s & 316.17s & 15.28 \\
MemGAS & 101.42s & 2.64s & 281.93s & 361.78s & 17.66 \\
\midrule
\method & \textbackslash & \textbf{0.74s} & \textbf{187.2s} & \textbf{187.96s} & \textbf{22.66} \\
\bottomrule
\end{tabular*}
\label{tab:efficiency}
\vspace{-2mm}
\end{table*}

\begin{table*}[t!]
\setlength\tabcolsep{0pt}  

\renewcommand{\arraystretch}{1.2} 
\small
\caption{Ablation Study on LoCoMo. 
(Full results can be found in Appendix Tab.~\ref{tab:appendix_ablation}.)}
\vspace{-2mm}
\centering
\begin{tabular*}{\linewidth}{@{\extracolsep{\fill}} l|ccccccc|c}
\toprule
\textbf{Model} &\textbf{4o-J}  &\textbf{F1} & \textbf{BLEU4}& \textbf{ROUGE1}&\textbf{ROUGE2}&\textbf{ROUGEL} & \textbf{BERTScore} & \textbf{\makecell{Avg.\\Tokens}}\\

\midrule
Baseline &40.33&15.76&2.77&16.08&7.75&15.10&84.70 & 2,348\\

w/ $TIR$ & 43.42 & 19.62 & 3.85 & 19.86 & 10.43 & 18.68 & 85.33 & 2,685 \\

w/ $TIR$ + $QDP$ & \textbf{48.84} & \textbf{22.66} & \textbf{5.33} & \textbf{22.92} & \textbf{12.04} & \textbf{21.68} & \textbf{85.94} & 1,403\\












\bottomrule
\end{tabular*}
\vspace{-4mm}
\label{tab:ablation}
\end{table*}

\begin{figure*}[t!]
  \centering

  \begin{minipage}[b]{0.48\textwidth}
    \centering
    \includegraphics[width=1.0\linewidth, height=4.2cm]{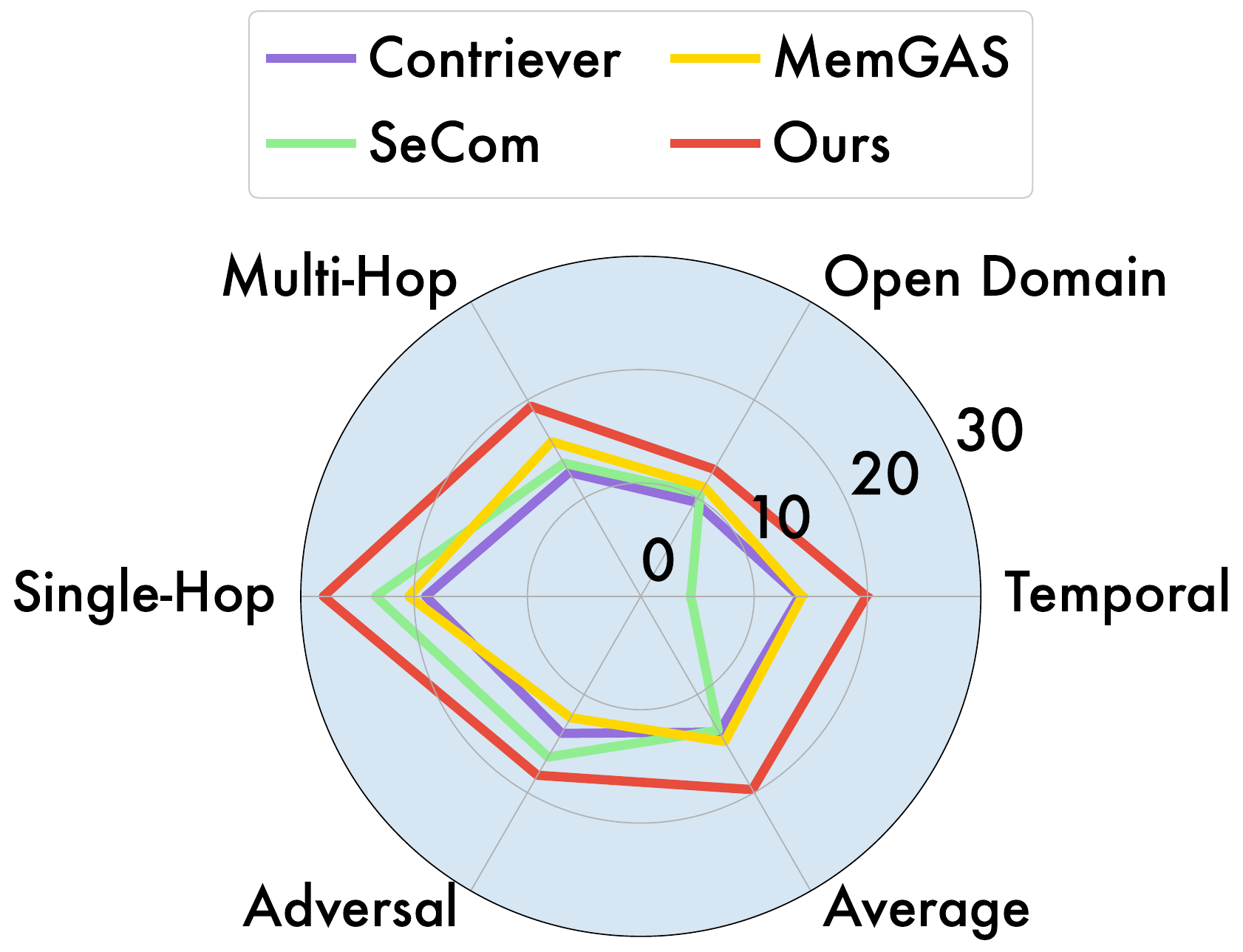}
  \end{minipage}
  \hfill
  \begin{minipage}[b]{0.48\textwidth}
    \centering
    \includegraphics[width=1.0\linewidth, height=4.2cm]{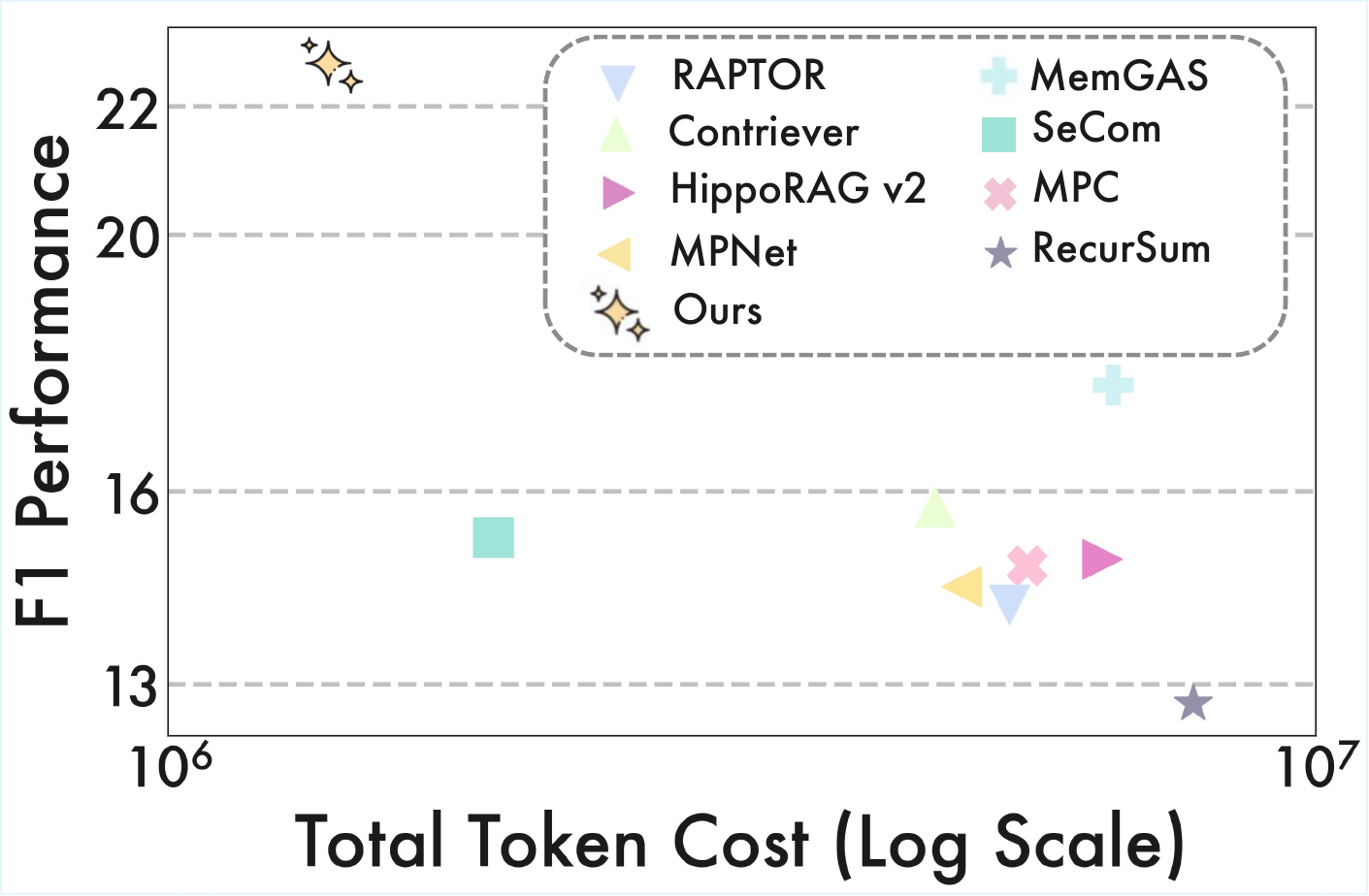}
  \end{minipage}
  
  \vspace{1mm}
  \begin{minipage}[t]{0.48\textwidth}
    \caption{Comparison of F1 scores across different query types in LoCoMo.}
    \label{fig:query}
  \end{minipage}
  \hfill
  \begin{minipage}[t]{0.48\textwidth}
    \caption{Trade-off between F1 score and total token consumption on LoCoMo.}
    \label{fig:token}
  \end{minipage}
  
  
  \begin{minipage}[b]{0.48\textwidth}
    \centering
    \includegraphics[width=1.0\linewidth, height=4.2cm]{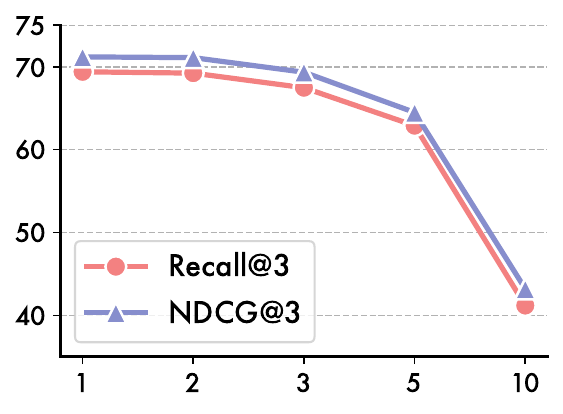}
  \end{minipage}
  \hfill
  \begin{minipage}[b]{0.48\textwidth}
    \centering
    \includegraphics[width=1.0\linewidth, height=4.2cm]{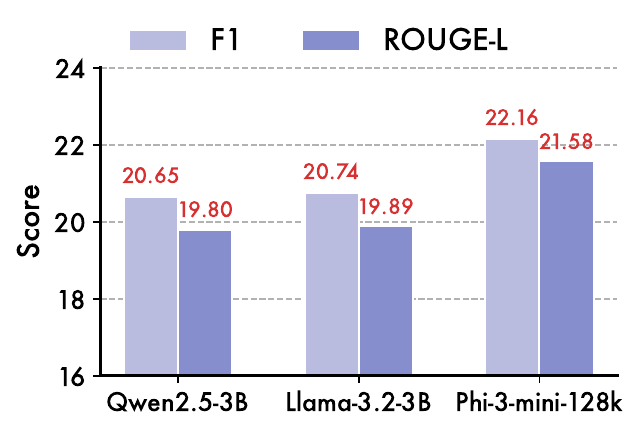} 
  \end{minipage}
  
  \vspace{1mm}
  \begin{minipage}[t]{0.48\textwidth}
    \caption{Retrieval performance when aggregating the top-$k$ relevant turns.}
    \label{fig:topk_turn}
  \end{minipage}
  \hfill
  \begin{minipage}[t]{0.48\textwidth}
    \caption{Generation performance when using different language models for pruning.}
    \label{fig:qdp_models}
  \end{minipage}
  \vspace{-4mm}
\end{figure*}

We evaluate the end-to-end computational overhead on LoCoMo dataset, covering time costs (Tab.~\ref{tab:efficiency}) and token consumption (Fig.~\ref{fig:token}).
\ding{182} \textit{Time Efficiency}: Tab.~\ref{tab:efficiency} demonstrates that \method reduces total execution time by 48.05\% compared to the strongest baseline. Traditional memory systems face a strict latency trade-off: they either incur heavy offline computation costs for structural maintenance, or suffer from severe online generation delays due to processing redundant historical contexts. \method effectively breaks this bottleneck. By operating directly on unstructured dialogue history, it mitigates offline construction overhead. 
Moreover, filtering conversational noise via QDP significantly accelerates online generation, confirming that our paradigm optimizes end-to-end efficiency without structural maintenance overhead.
\ding{183} \textit{Token Efficiency}: 
As illustrated in Fig.~\ref{fig:token}, \method~ achieves an outstanding trade-off between F1 performance and token consumption. 
Baselines typically incur heavy token overhead by requiring iterative LLM inferences for offline memory construction or directly processing noisy top-$k$ contexts. By bypassing offline construction and explicitly pruning redundant contexts, \method reduces overall token consumption by 3--5$\times$ while increasing response accuracy. 
This exposes a fundamental inefficiency in existing memory paradigms: they allocate substantial computational budget to processing conversational filler, rather than focusing strictly on sparse query-relevant signals.

\subsection{Ablation \& Mechanism Analysis}

\noindent \textit{Ablation Study.} 
Tab.~\ref{tab:ablation} shows that the integration of TIR and QDP is essential to mitigate the two facets of the \textit{Signal Sparsity Effect}. 
At the retrieval stage, TIR successfully overcomes Decisive Evidence Sparsity by directly isolating high-evidence signals within the manifold, resulting in an initial improvement of F1 to 19.62 on LoCoMo. 
However, feeding these raw retrieved sessions directly into the generator degrades response quality due to conversational noise.
At the generation stage, \method mitigates \textit{dual-level redundancy} by pruning this redundant context via QDP, which further boosts the F1 score to 22.66 while sharply reducing token consumption from 2,685 to 1,403. This suggests that improving signal density can serve as a highly effective alternative to increasing architectural complexity.
Complete ablation results across varying query settings are provided in Appendix~\ref{appendix_more_exp_ablation}.

\noindent \textit{TIR Mechanism.} We determine a candidate session's relevance score by aggregating the query-turn similarities of its top-$k$ most relevant turns. As illustrated in Fig.~\ref{fig:topk_turn}, increasing the number of retained turns ($k$) sharply degrades overall performance, because the aggregation of extra turns introduces conversational noise that dilutes the decisive matching signal.

\noindent \textit{QDP Mechanism.} To examine whether QDP relies on specific generator architectures, we evaluate the pruning effectiveness using lightweight models including Qwen2.5-3B, Llama-3.2-3B, and Phi-3-mini-128k.
As shown in Fig.~\ref{fig:qdp_models}, \method maintains robust performance across all pruners, demonstrating that the pruning mechanism is compatible with various lightweight backbones and does not require architectural alignment with the generator.

\section{Related Work}

\subsection{The Representation Ability for Better Retrieval.}
Long-term conversational memory inherently requires segmentation strategies to manage growing contexts~\citep{locomo,longmemeval,secom}. Existing methods (categorized in Appendix~\ref{appendix_baseline}, Tab.~\ref{tab:comparison}) predominantly rely on \textit{single-granularity} representations, utilizing discrete conversational units (e.g., \textit{turn-level}~\citep{yuan2024personal} and \textit{session-level}~\citep{li2025hello, memochat, MPC}), abstractive compressions (e.g., \textit{summary-level}~\citep{zhong2024memorybank,chen2025compress,memochat}, \textit{keyword-level}~\citep{xu2025amem, chhikara2025mem0,zhou2025simple, ke2025flexibly} or semantic boundary \textit{topic-level}~\citep{secom,fang2025lightmem,fountas2025human}). However, these approaches face a fundamental limitation: they either fragment episodic context or induce semantic dilution, losing decisive evidence during retrieval.
To address this, \textit{Multi-granularity} paradigm~\citep{memgas,jiang2026magma,li2026timem} fuses representations across multiple granularities yet introduces redundancy.
Unlike these paradigms, our retrieval mechanism directly retrieves the single most relevant turn, preserving the core matching signal from raw dialogue, effectively avoiding context dilution without compromising precision.

\subsection{The Information Feature for Better Responses.}
Existing memory systems~\citep{xu2025amem,packer2024memgpt,chhikara2025mem0} 
primarily focus on memory construction and retrieval, ignoring noise within the retrieved contexts.
Post-retrieval reranking methods reorder retrieval contexts with heuristic rules~\citep{liu2025rcr} or parametric learning ~\cite{tan2025memotime,du2025memguide}, however, they fail to remove redundant information from the retrieved contexts.
Constraint-based approaches~\citep{rasmussen2025zep,han2025rag} use fixed rules to filter retrieved contexts based on temporal/logical conflicts, but they depend on hand-crafted metadata and cannot adapt to different queries.
Multi-granularity~\citep{memgas,li2026timem} enrich the generation phase by appending vary representations(e.g., keywords, summaries) from the retrieved contexts, introducing representational redundancy and computational overhead.
In contrast, our QDP filters the retrieved contexts, ensuring robust generation while reducing noise.
\section{Conclusion, Limitations and Future Work}
In this paper, we proposed \method~, a nonparametric memory for conversational agent.
By designing TIR that scores each past session by its single most relevant turn and QDP that filters redundant retrieved context, our method effectively preserves sparse critical information while eliminating noise.
Extensive experiments demonstrate its effectiveness, universality, and efficiency. 
However, \method~'s passive, unstructured paradigm limits the agent's capacity for proactive self-evolution and continuous knowledge consolidation. Furthermore, the reactive nature of online query-driven pruning introduces sequential inference latency. Future research will focus on transitioning this framework into a proactive, self-evolving memory system capable of autonomous, offline knowledge restructuring.

\clearpage

\section*{Ethics Statement}
This paper presents work whose goal is to advance the field of language models. There are many potential societal consequences of our work, none of which we feel must be specifically highlighted here.

\bibliography{colm2026_conference}
\bibliographystyle{colm2026_conference}

\appendix

\onecolumn
\begin{center}
    \Large{\sc\Huge Supplementary Material\\\small Back to Basics: Let Conversational Agents Remember \\with Just Retrieval and Generation}\\
\end{center}
\vskip 4mm
\startcontents[sections]
\vbox{\sc\Large Table of Contents}
\vspace{5mm}
\hrule height .8pt
\vspace{-2mm}
\printcontents[sections]{}{1}{\setcounter{tocdepth}{2}}
\vspace{4mm}
\hrule height .8pt
\vskip 10mm

\section{Experimental Details}\label{appendix_setup}

\subsection{Datasets Details}
\label{appendix_datasets}
\begin{table*}[htbp]
\centering
\caption{Detailed statistics of all evaluated datasets. 
The term "Avg." (e.g., Avg. Queries) denotes the average number corresponding to each conversation.}
\resizebox{\linewidth}{!}{
\begin{tabular*}{\linewidth}{@{\extracolsep{\fill}} l|c|c|c|c }
\toprule
\textbf{Dataset} & {\small \textbf{LoCoMo}} & {\small \textbf{Long-MT-Bench+}} & {\small \textbf{LongMemEval-s}} & {\small \textbf{LongMemEval-m}} \\
\midrule
Conversation Subject & User-User & User-AI & User-AI & User-AI \\
Session Dates & \gcmark & \rxmark & \gcmark & \gcmark \\
Retrieval Ground-Truth & \gcmark & \rxmark & \gcmark & \gcmark \\
QA Ground-Truth & \gcmark & \gcmark & \gcmark & \gcmark \\
Total Conversations & 10 & 11 & 500 & 500 \\
Avg. Sessions & 27.2 & 4.9 & 47.7 & 475.3 \\
Avg. Queries & 198.6 & 26.2 & 1.0 & 1.0 \\
Avg. Turns & 301.1 & 65.5 & 248.7 & 2466.4 \\
Avg. Tokens & 20,756.2 & 19,287.5 & 104,315.2 & 1,030,374.7 \\
\bottomrule
\end{tabular*}
}
\label{tab:dataset_statistics}
\end{table*}

\textbf{LoCoMo}~\citep{locomo}
is designed to evaluate long-term memory in long-context LLMs and RAG systems in multi-turn question-answering tasks.
In this study, we employ its publicly available subset, which contains 10 high-quality, long conversations with 27.2 sessions and 20,756.2 tokens on average.
The evaluation tasks consist of 1,986 queries covering five problem types: (i) single-hop retrieval, which sources answers from a specific isolated session; (ii) multi-hop retrieval, which necessitates synthesizing details scattered across multiple sessions; (iii) temporal reasoning, demanding the deduction of chronological sequences and time intervals; (iv) open-domain knowledge, which requires grounding speaker information in external commonsense or world knowledge ; and (v) adversarial, evaluating the model's ability to detect and refuse to answer misleading or unanswerable prompts.

\textbf{Long-MT-Bench+}~\citep{secom} is an enhanced long-dialogue benchmark reconstructed from MT-Bench+~\citep{memochat}. It addresses scarce QA pairs and short dialogues by merging five consecutive sessions into a long-form conversation. The dataset contains 11 conversations, averaging 4.9 sessions and approximately 19287.5 tokens. Unlike LoCoMo, it focuses exclusively on user-AI interactions and does not provide session timestamps or retrieval ground truth.

\textbf{LongMemEval-s}~\citep{longmemeval} follows the needle-in-a-haystack paradigm, focusing on dynamically tracking and updating information in task-oriented conversations. The benchmark includes 500 manually designed questions, each accompanied by a chat history of about 47.7 sessions and averaging 104,315.2 tokens. 
The queries are systematically categorized to evaluate diverse memory dimensions: (i) single session user, recalling specific details provided by the user within an isolated interaction; (ii) single session assistant, retrieving information previously generated by the agent in a single session; (iii) single session preference, personalizing responses based on user traits shared locally; (iv) multi-session, synthesizing dispersed evidence across multiple sessions to address complex questions; (v) temporal reasoning, resolving chronological dependencies, including explicit dates and inferred time references and (vi) knowledge update, tracking dynamically evolving user states and reasoning with overridden personal information over time.

\textbf{LongMemEval-m}~\citep{longmemeval} is an extended version of LongMemEval-s, scaling each dialogue history to an average of 500 sessions and 1030374.7 tokens. 
It is designed to detect lost-in-the-middle phenomenon in long-context models and to assess their ability to maintain temporal consistency and perform causal reasoning over extended contexts.
Both versions challenge AI systems to process lengthy user-assistant interactions, maintain dynamically evolving memory, and ensure consistency throughout conversational history.

\subsection{Baseline Details}
\label{appendix_baseline}

In this appendix, we provide detailed descriptions of the advanced memory methods and different RAG systems used in our default evaluation.

\vspace{0.5em}
\begin{itemize}[noitemsep, topsep=8pt, partopsep=0pt, leftmargin=6mm,parsep=8pt]
\setlength\itemsep{0mm}

\begin{table*}[t!]
\centering
\renewcommand{\arraystretch}{1.8}
\caption{Comparison of various retrieval granularities. \textbf{Efficiency} is measured by Tokens (generation consumption) and Latency. \textbf{Simplicity} is quantified by direct inference over raw dialogue history and the avoidance of representational redundancy (e.g., multi-granularity clustering).\High: High, \Medium: Moderate, \Low: Low. $*$ indicates a specific variant that isolates the decisive turn to mitigate context dilution.}
\label{tab:comparison}
\resizebox{\linewidth}{!}{
\begin{tabular}{ccccccc}
\toprule
\textbf{Category} & \textbf{Example Works} & \textbf{Operation} & \textbf{Effectiveness} & \textbf{Simplicity} & \textbf{Tokens} & \textbf{Latency} \\
\cmidrule(lr){1-1}\cmidrule(lr){2-7}

\makecell[c]{Keyword-Level} & A-Mem~\citeyearpar{xu2025amem}, H-MEM~\citeyearpar{sun2025hierarchical} & Extract turn/session details & \Low & \Medium & \Medium & \Medium \\
\hline

\makecell[c]{Turn-Level} & Conditional Memory~\citeyearpar{yuan2024personal}

& Raw Turn Retrieval & \Low & \High & \High & \High \\
\hline
\makecell[c]{Session-Level} & MemoChat~\citeyearpar{memochat}, LD-Agent~\citeyearpar{li2025hello}& Raw Session Retrieval & \Low & \High & \High & \High \\
\hline
\makecell[c]{Summary-Level} & Memorybank~\citeyearpar{zhong2024memorybank}, Recursum~\citeyearpar{RecurSum} & Recurrent summary & \Low & \Medium & \Medium & \Medium \\
\hline
\makecell[c]{Topic-Level} & Secom~\citeyearpar{secom}, LightMem~\citeyearpar{fang2025lightmem} & Semantic segmentation & \Medium & \Medium & \Medium & \Medium \\

\hline
\makecell[c]{Multi-Granularity} & MemGAS~\citeyearpar{memgas}, TiMem~\citeyearpar{li2026timem} & Adaptive Fusion & \High & \Low & \Low & \Low \\

\hline
\rowcolor{gray!8} $Turn-Level^{*}$ & \textbf{\method (Ours)} & Mitigate Sparsity & \High & \High & \High & \High \\
\bottomrule
\end{tabular}
}
\vspace{-4mm}
\end{table*}

\item \textit{Full History} is a retrieval-free baseline that feeds the most recent 128k tokens of dialogue history directly into the LLM’s context window, allowing the model to process the complete, unedited history.

\item \textit{MPNet} ~\citep{song2020mpnet} 
is a pre-training approach for LLMs that combines the strengths of masked language modeling from BERT~\citep{liu2019roberta} and permuted language modeling from Xlnet~\citep{yang2019xlnet}, while addressing their key limitations. It captures dependencies among predicted tokens through permutation-based objectives and incorporates auxiliary position information to align pre-training more closely with downstream tasks, enabling better utilization of full sentence context. This method particularly excels in natural language understanding and downstream performance across a range of tasks.
\url{https://github.com/microsoft/MPNet}

\item \textit{Contriever}~\citep{contriever} 
is an unsupervised dense retrieval model trained via contrastive learning, enabling effective representation of queries and documents without labeled data. It demonstrates strong capabilities in zero-shot retrieval across diverse domains, as well as in-domain performance when further fine-tuned, and particularly excels in multilingual and cross-lingual retrieval settings, including challenging scenarios involving different scripts.
\url{https://github.com/facebookresearch/contriever}

\item \textit{MPC}~\citep{MPC}
is a modular approach to building conversational agents that leverages pre-trained large language models as independent components to achieve long-term consistency and flexibility. It employs techniques such as few-shot prompting, chain-of-thought reasoning, and external memory, enabling the creation of consistent and engaging chatbots without requiring model fine-tuning.
\url{https://github.com/krafton-ai/MPC}

\item \textit{RAPTOR}~\citep{raptor}
is a retrieval-augmented framework that recursively embeds, clusters, and summarizes text chunks to build a multi-layer tree structure with increasing levels of abstraction from bottom-up summaries. At inference time, it retrieves relevant information across different abstraction levels, enabling effective integration of context from long documents. RAPTOR performs strongly on long-context QA tasks involving single-hop and multi-hop reasoning.
\url{https://github.com/parthsarthi03/raptor}

\item \textit{RecurSum}~\citep{RecurSum}
is a recursive summarization approach that enhances long-term dialogue capabilities in large language models by progressively generating updated memories. It begins by having the LLM memorize short dialogue contexts, then iteratively produces new summaries incorporating prior memories and additional conversation segments, enabling the chatbot to generate responses grounded in the most recent consolidated memory. This method particularly excels in maintaining consistency across extended conversations.
\url{https://github.com/qingyue2014/Rsum}

\item \textit{HippoRAG 2}~\citep{hipporag2}
is a retrieval-augmented framework that advances long-term memory capabilities in LLMs by leveraging an enhanced Personalized PageRank algorithm, incorporating deeper passage integration and efficient online LLM utilization. It excels in factual knowledge recall, sense-making, and associative memory tasks.
\url{https://github.com/OSU-NLP-Group/HippoRAG}

\item \textit{A-Mem}~\citep{xu2025amem}
is an agentic memory system for LLM agents that dynamically organizes memories by drawing inspiration from the Zettelkasten method to build interconnected knowledge networks. When adding a new memory, a-mem generates structured notes with contextual descriptions, keywords, and tags, then identifies connections with historical memories to establish links, while enabling memory evolution through updates to existing representations as new information is integrated.
\url{https://github.com/WujiangXu/A-mem}

\item \textit{SeCom}~\citep{secom}
is a memory construction framework that builds a segment-level memory bank using a conversation segmentation model to divide long-term conversations into topically coherent units, while applying LLMLingua-2~\citep{pan2024llmlingua} compression to denoise memory entries and improve retrieval performance.         
\url{https://github.com/microsoft/SeCom}

\item \textit{MemGAS}~\citep{memgas}
is a memory consolidation framework that organizes memories into units of varying granularity and uses Gaussian Mixture Models to cluster and associate new memories with existing ones. An entropy-based router dynamically selects the most appropriate granularity for each query by assessing relevance distributions, ensuring a balance between information completeness and minimal noise.
\url{https://github.com/quqxui/MemGAS}

\end{itemize}

\subsection{Implementation Details}
\label{appendix_details}
Across all tasks, we utilize \textit{gpt-4o-mini-2024-07-18} as the default generator. To ensure strict experimental reproducibility and eliminate generative randomness, we set the LLM temperature to 0 and cap the maximum response length at 4000 tokens. 
All evaluations are conducted on a Linux server featuring an AMD EPYC 7763 128-Core Processor with 256GB of RAM, accelerated by 4 $\times$ NVIDIA RTX A6000 GPUs, each with 48GB memory. 
\section{More Experimental Analysis}~\label{appendix_more_exp}
\vspace{-3mm}

\subsection{Effectiveness Analysis}~\label{appendix_more_exp_effectiveness}

We provide complete retrieval performance across multiple metrics on three long-term conversational benchmarks, as presented in Tab.~\ref{tab:appendix_retrieval_whole}. Note that LongMTBench+ is excluded from this evaluation because it lacks retrieval ground truth.
\begin{table*}[t!]
\setlength\tabcolsep{0pt}  
\small
\centering
\caption{Retrieval Performance. Contriever is used as the default retriever for all methods except MPNet.}

\begin{tabular*}{\linewidth}{@{\extracolsep{\fill}} l|cccccc|c }
\toprule
\textbf{Model} &\textbf{Recall@3 }&\textbf{NDCG@3}&\textbf{Recall@5} &\textbf{NDCG@5}&\textbf{Recall@10} & \textbf{NDCG@10} & \textbf{Time}\\

\midrule
\multicolumn{8}{c}{\cellcolor{cyan!5}\textit{\textbf{LoCoMo}}} \\
\midrule
MPNet ~\citeyearpar{song2020mpnet}& 45.92 & 47.68 & 53.98 & 51.76 & 68.58 & 56.85 & 1.42s\\

Contriever ~\citeyearpar{contriever} &49.95& 52.15 &58.31& 56.29& 71.80& 60.91 & 1.24s\\

MPC ~\citeyearpar{MPC} & 49.50  & 51.47  & 57.45  & 55.53  & 71.85  & 60.47 & 0.81s \\

RecurSum ~\citeyearpar{RecurSum} &47.23&48.99&59.01&54.58&74.97&60.07 & 1.26s\\

SeCom ~\citeyearpar{secom} & {52.97}& {54.03}& {64.05}& {59.37}& {78.25}& {64.28}& 2.17s\\
HippoRAG 2~\citeyearpar{hipporag2} &56.60&58.37&65.06&62.50&78.05&66.79& 3.70s\\
MemGAS~~\citeyearpar{memgas}  &\underline{56.85}&  \underline{59.22}&  \underline{67.98}& \underline{64.25}&   \underline{81.22}& \underline{68.71}  & 2.64s \\
\method & \textbf{69.39}  & \textbf{71.19}  & \textbf{76.33} &  \textbf{74.58} &  \textbf{86.46}  & \textbf{77.99} & 0.74s\\  
\midrule
\multicolumn{8}{c}{\cellcolor{cyan!5}\textit{\textbf{LongMemEval-s}}} \\
\midrule

MPNet ~\citeyearpar{song2020mpnet}&66.60&75.86&77.02&78.72&85.96&81.07& 0.47s\\

Contriever ~\citeyearpar{contriever}&71.06 & 79.84 & 81.28 & 82.56 & 90.00 & 84.38& 0.39s\\

MPC ~\citeyearpar{MPC} & 75.32 &  87.47  & 75.96 &  86.79  & 77.23 &  87.17 & 0.71s\\

RecurSum ~\citeyearpar{RecurSum} &60.00&70.90&68.09&73.27&80.00&76.59& 0.57s\\

SeCom ~\citeyearpar{secom}
& 71.06 & 80.88 & 80.43 & 83.08 & 89.15 & 85.11 & 0.79s\\ 

HippoRAG 2~\citeyearpar{hipporag2} & 75.53 & 85.44 &84.68&87.32&91.28&88.73& 21.46s\\

MemGAS~\citeyearpar{memgas}  & \textbf{78.51}  & \textbf{86.83} &  \textbf{88.30}  &  \textbf{88.86} &  \underline{92.55}   & \underline{89.79}  & 2.19s\\
\method & \underline{77.23}  & \underline{85.77}  & \underline{87.02} &  \underline{88.33} &  \textbf{94.26}  & \textbf{89.81} & 0.25s \\

\midrule
\multicolumn{8}{c}{\cellcolor{cyan!5}\textit{\textbf{LongMemEval-m}}} \\
\midrule
MPNet ~\citeyearpar{song2020mpnet}
& 38.09  & 49.91 &  47.23  & 53.81 &  61.70 &   57.51 & 2.16s\\

Contriever ~\citeyearpar{contriever}
& 45.32  & 57.18  & \underline{54.26} &  60.42  & 66.81 &  63.63 & 1.78s \\

MPC ~\citeyearpar{MPC} &35.96&47.51&42.55&50.36&54.26&53.88& 2.34s\\


SeCom ~\citeyearpar{secom} 
& \underline{51.49}  & \underline{63.31} &  \textbf{63.62} &  \underline{67.14}  & 68.72  & 68.68 & 3.10s\\
MemGAS~\citeyearpar{memgas} &51.06&61.36& \textbf{63.62}&66.07&\textbf{77.02}&\underline{69.46}& 8.88s\\

\method & \textbf{52.77} & \textbf{64.61}  & \textbf{63.62} &  \textbf{68.23}  & \underline{75.74}   & \textbf{71.42} & 2.08s\\
\bottomrule
\end{tabular*}

\label{tab:appendix_retrieval_whole}
\end{table*}

\subsection{Universality Analysis}~\label{appendix_more_exp_university}
\vspace{-2mm}

\begin{figure*}[t!]
  \centering
  \subfloat{%
    \includegraphics[width=0.32\textwidth]{figure/exp/every_type/locomo_con.pdf}  
    \label{fig:intro_b}
  }
  \hfill
  \subfloat{%
    \includegraphics[width=0.32\textwidth]{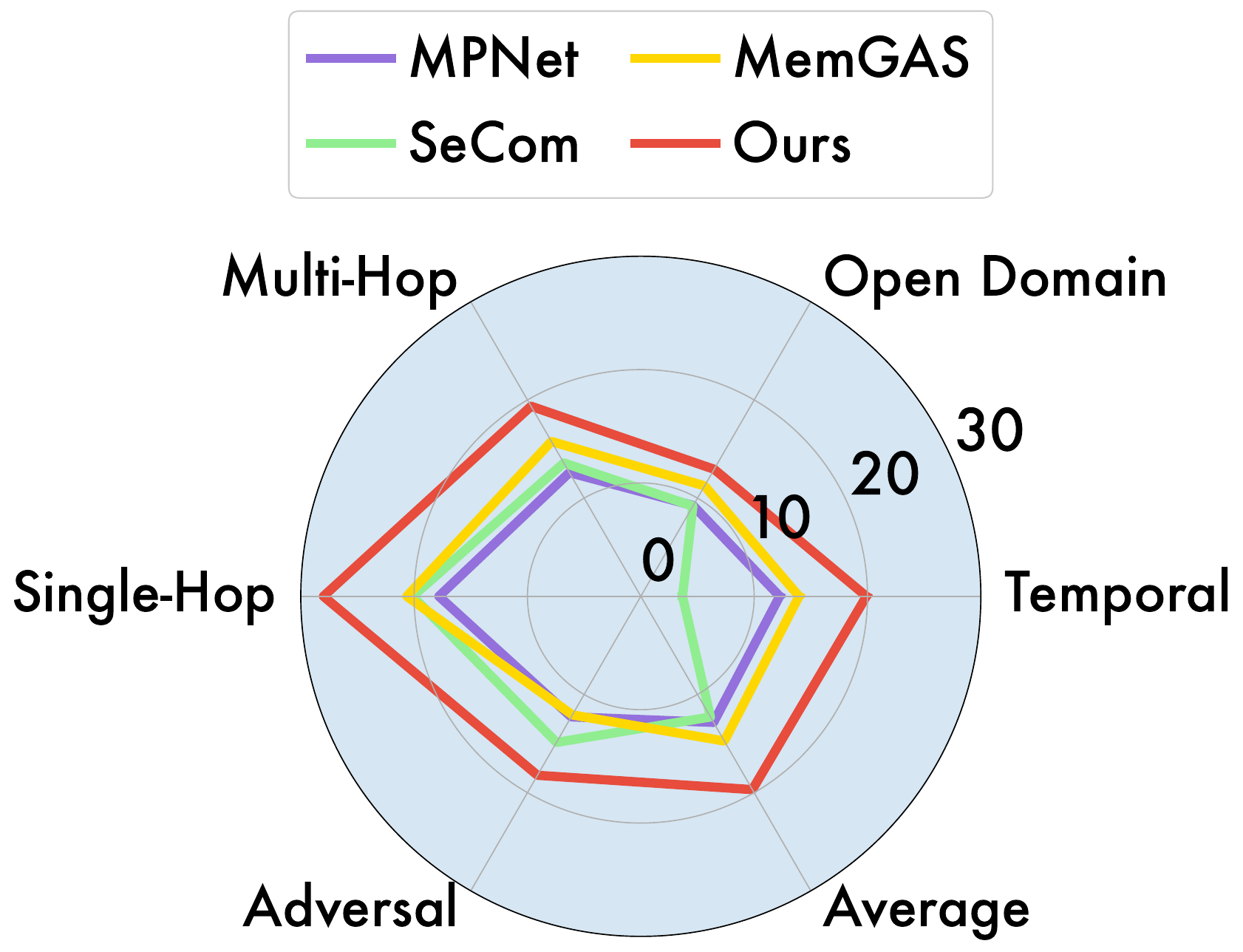}  
    \label{fig:intro_b}
  }
  \hfill
  \subfloat{%
    \includegraphics[width=0.32\textwidth]{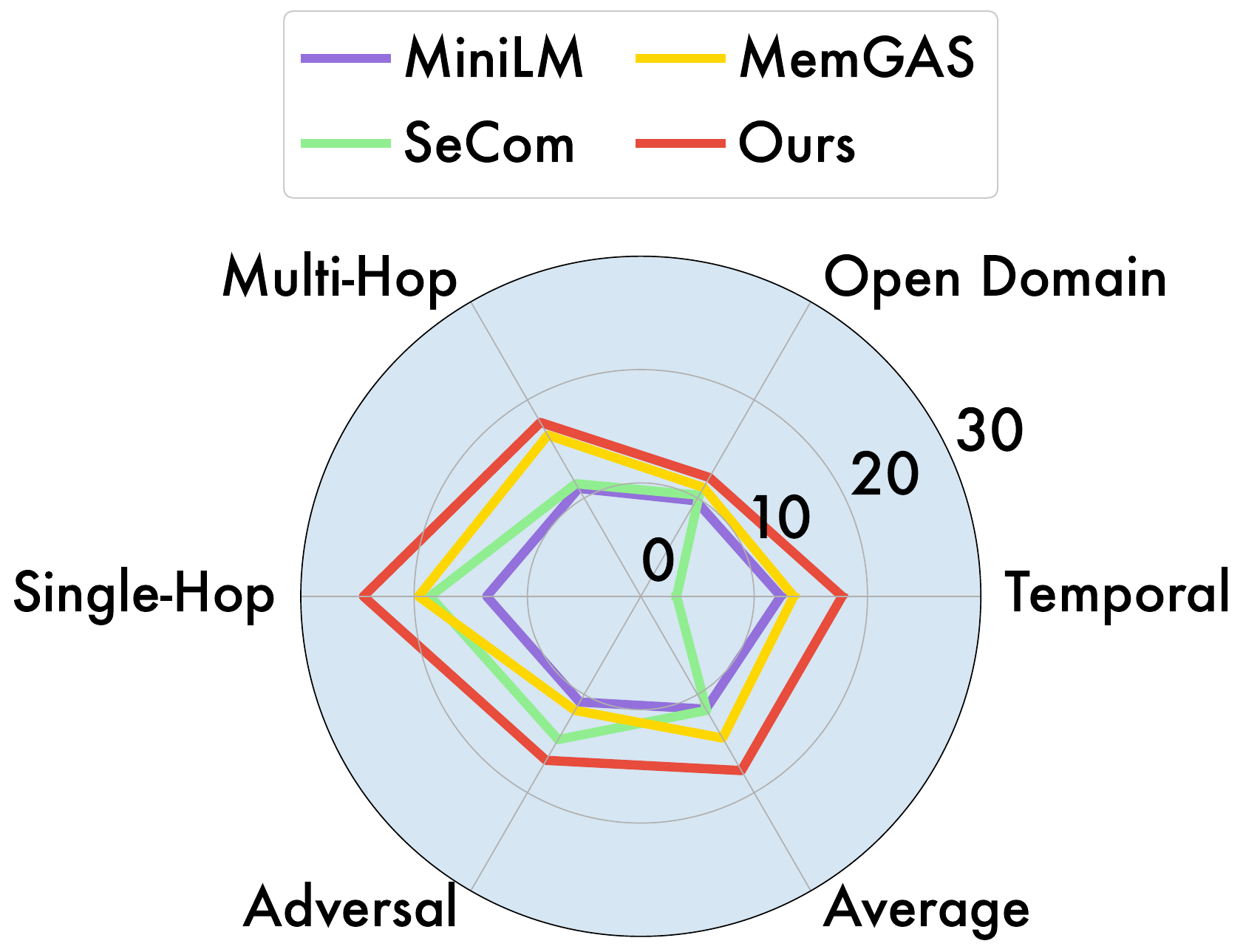}   
    \label{fig:intro_c}
  }
  \vspace{-2mm}  
  \caption{Comparison of F1 scores across different query types in LoCoMo using \textit{gpt-4o-mini-2024-07-18} as the generator with Contriever(left), MPNet (middle), and MiniLM (right)
  }
  \label{fig:whole_query_retriever}
  \vspace{-4mm}
\end{figure*}

We provide more experimental results for the universality analysis.
Specifically, Tab.~\ref{tab:diff_gen} presents the QA generation performance across different LLM backbones. Tab.~\ref{tab:diff_retriever} and Tab.~\ref{tab:diff_gen_retriever} detail the intermediate retrieval and final QA performance across different retrievers, respectively. Fig~\ref{fig:whole_query_retriever} illustrates the generation performance of all queries with different retrievers. 
We observe similar analysis as in the main text.

\begin{figure*}[t!]
  \centering

  \subfloat[LoCoMo, Contriever]{%
    \includegraphics[width=0.32\textwidth]{ 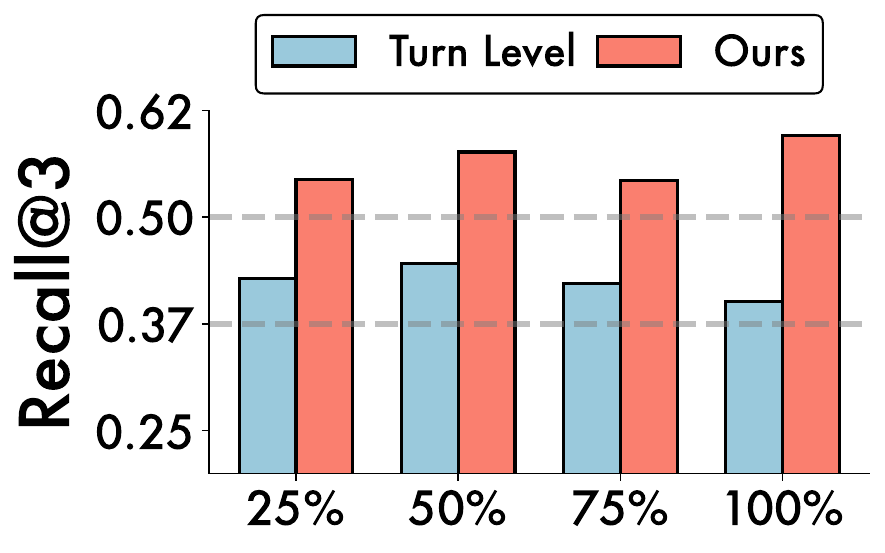}
    \label{fig:row1_col1}
  }
  \hfill
  \subfloat[LoCoMo,MPNet]{%
    \includegraphics[width=0.32\textwidth]{ 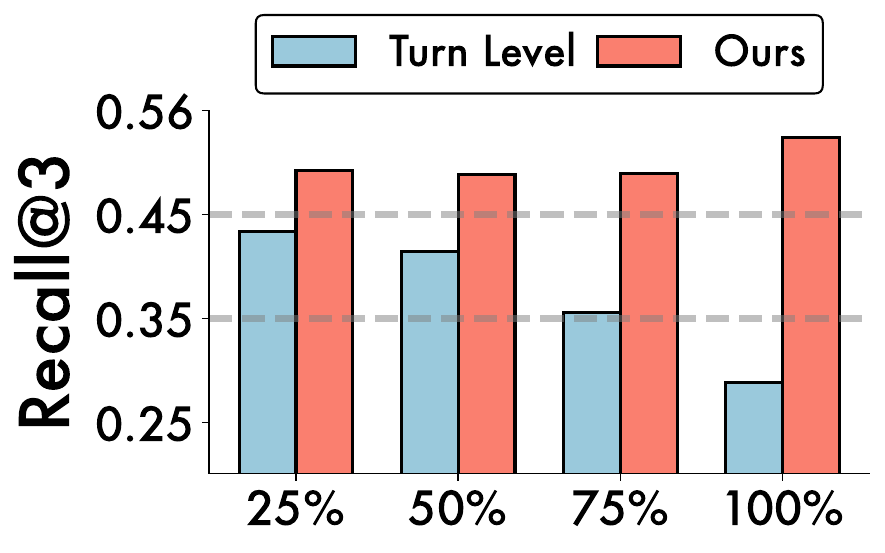}
    \label{fig:row1_col2}
  }
  \hfill
  \subfloat[LoCoMo,MiniLM]{%
    \includegraphics[width=0.32\textwidth]{ 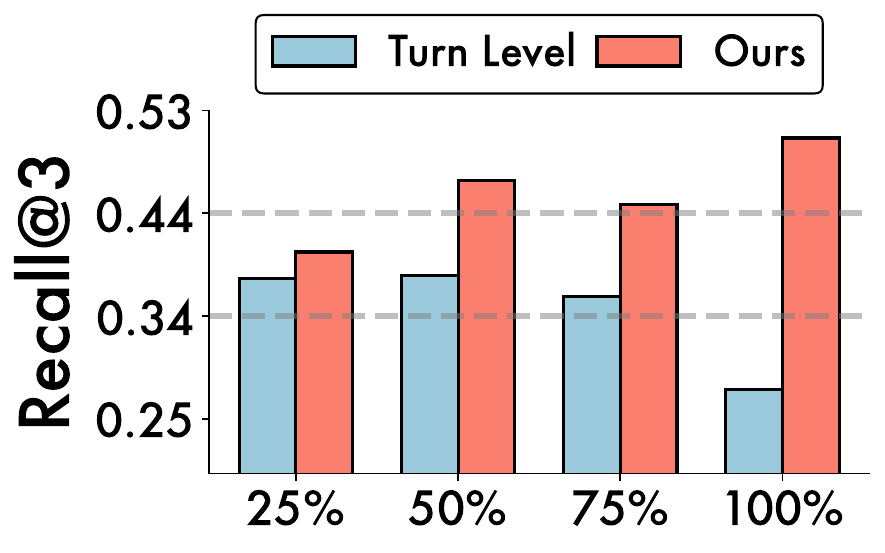}
    \label{fig:row1_col3}
  }
  \\ \vspace{1mm} 

  \subfloat[LongMemEval-s, Contriever]{%
    \includegraphics[width=0.32\textwidth]{ 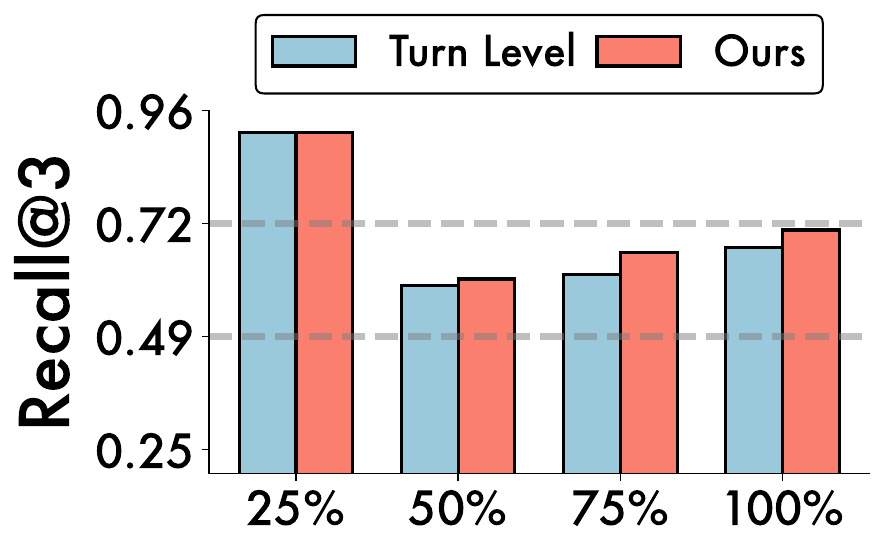}
    \label{fig:row2_col1}
  }
  \hfill
  \subfloat[LongMemEval-s,MPNet]{%
    \includegraphics[width=0.32\textwidth]{ 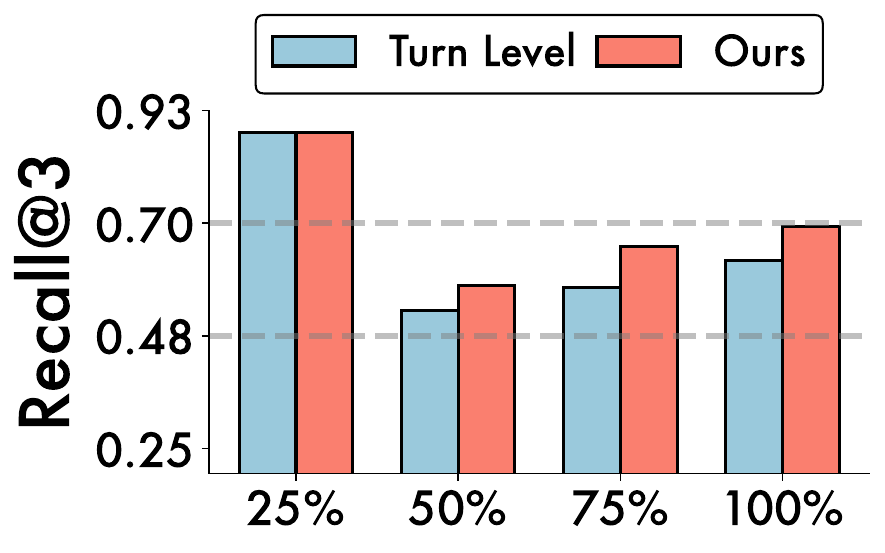}
    \label{fig:row2_col2}
  }
  \hfill
  \subfloat[LongMemEval-s,MiniLM]{%
    \includegraphics[width=0.32\textwidth]{ 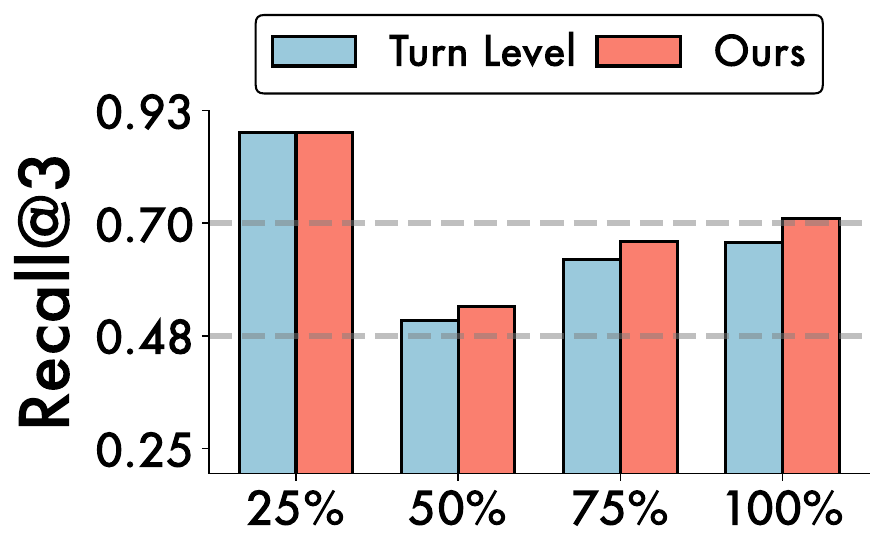}
    \label{fig:row2_col3}
  }
  \\ \vspace{1mm} 
  \subfloat[LongMemEval-m, Contriever]{%
    \includegraphics[width=0.32\textwidth]{ 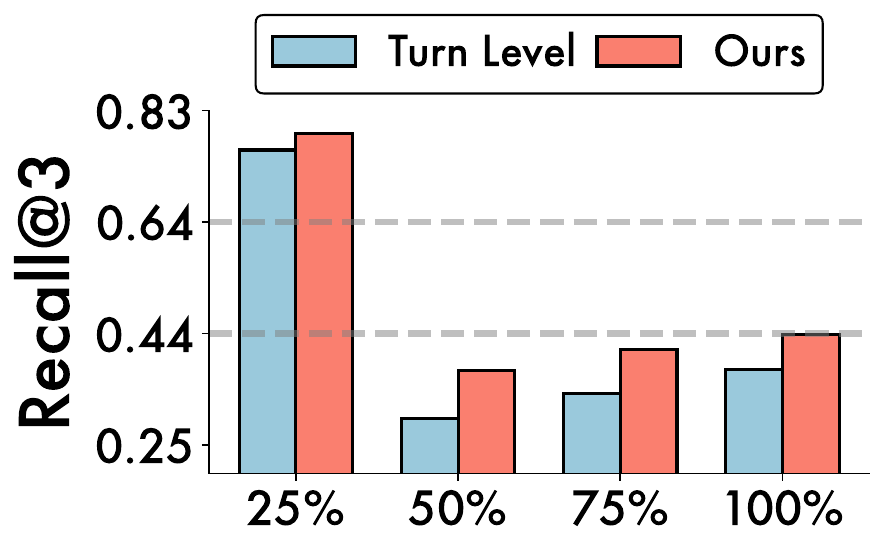}
    \label{fig:row3_col1}
  }
  \hfill
  \subfloat[LongMemEval-m,MPNet]{%
    \includegraphics[width=0.32\textwidth]{ 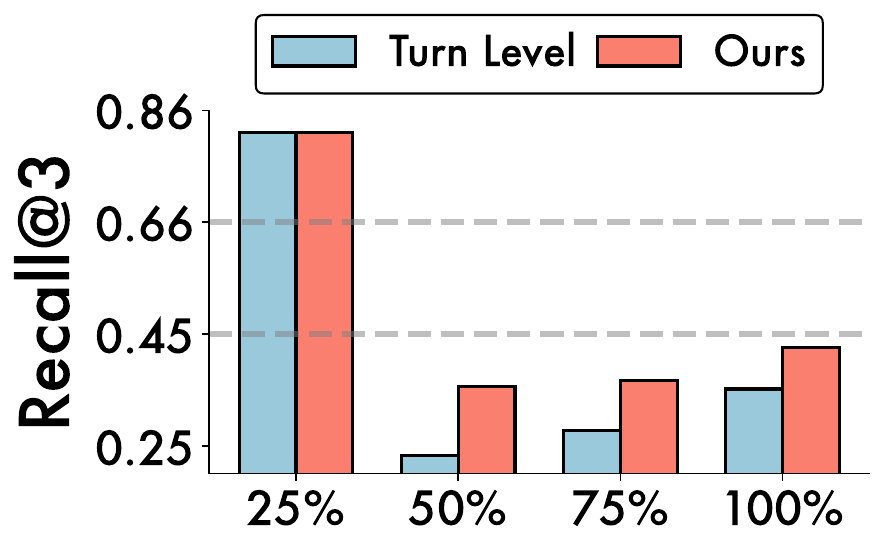}
    \label{fig:row3_col2}
  }
  \hfill
  \subfloat[LongMemEval-m, MiniLM]{%
    \includegraphics[width=0.32\textwidth]{ 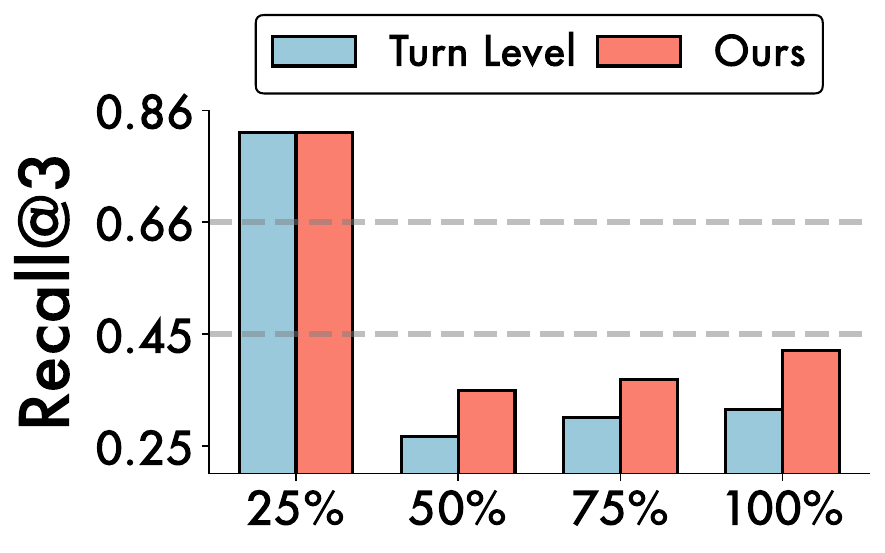}
    \label{fig:row3_col3}
  }
  
  \vspace{-2mm}
  \caption{Recall@3 performance comparison across LoCoMo, LongMemEval-s, LongMemEval-m using Contriever,MPNet,MiniLM. We rank all sessions by their total turn counts and group them into four equal percentiles (0--25\%, 25\%--50\%, 50\%--75\%, 75\%--100\%).} 
  \label{fig:appendix_bucket_9grid}
  \vspace{-4mm}
\end{figure*}

To further substantiate \textit{Finding I} and the initial trends observed in Fig~\ref{fig:method_1}(c) of the main text, Fig~\ref{fig:appendix_bucket_9grid} illustrates the retrieval performance across different session length percentiles using three datasets and three dense retrievers. 
The results reveal a consistent phenomenon: as session length increases, the performance of the Turn-level baseline (which relies on mean-pooling) degrades significantly, particularly in the longest quartile (75\%-100\%). This confirms that accumulating extended conversational history inherently dilutes the core matching signals. Conversely, our proposed retrieval mechanism maintains robust performance across all length buckets. By retaining the single most relevant turn, our approach effectively neutralizes the context dilution effect, demonstrating strong generalizability across varying dialogue lengths and retriever architectures.
\begin{table*}[t!]
\setlength\tabcolsep{0pt}  

\small
\caption{QA Performance Comparison of Different Generators on LoCoMo with Contriever.}
\centering
\begin{tabular*}{\linewidth}{@{\extracolsep{\fill}} l|ccccccc}
\toprule
\textbf{Model} &\textbf{4o-J}  &\textbf{F1} & \textbf{BLEU4}& \textbf{Rouge1}&\textbf{Rouge2}&\textbf{RougeL} & \textbf{BERTScore} \\

\midrule
\multicolumn{8}{c}{\cellcolor{cyan!5}\textit{ \textbf{GPT-4o-mini-2024-07-18}}} \\
\midrule
Contriever ~\citeyearpar{contriever}&40.33&15.76&2.77&16.08&7.75&15.10&84.70 \\

SeCom ~\citeyearpar{secom} & 43.45&15.28& 3.12 & 17.16 & 8.52 & 16.01 & 84.84 \\

MemGAS~\citeyearpar{memgas}  &41.07 & 17.66 &  3.61 & 18.00& 8.93 & 16.99 & 85.13 \\
\method & \textbf{48.84} & \textbf{22.66} & \textbf{5.33} & \textbf{22.92} & \textbf{12.04} & \textbf{21.68} & \textbf{85.94} \\

\midrule
\multicolumn{8}{c}{\cellcolor{cyan!5}\textit{ \textbf{Llama3.1-8b-instruct}}} \\
\midrule
Contriever ~\citeyearpar{contriever}& 37.97 & 12.91 & 1.70 & 13.19 & 6.38 & 12.41 & 83.90 \\
SeCom ~\citeyearpar{secom} & 40.58 & 14.78 & 2.06 & 15.06 & 7.55 & 14.08 & 84.27 \\
MemGAS~\citeyearpar{memgas}& 34.69 & 14.24 & 1.66 & 14.64 & 7.20 & 13.83 & 84.14 \\
\method &\textbf{44.66}& \textbf{19.90} & \textbf{3.11} & \textbf{20.03} & \textbf{10.84} & \textbf{19.07} & \textbf{85.19} \\

\midrule
\multicolumn{8}{c}{\cellcolor{cyan!5}\textit{ \textbf{gpt-5.4-mini}}} \\
\midrule
Contriever ~\citeyearpar{contriever} & 46.68 & 21.44 & 3.79 & 22.33 & 10.84 & 21.01 & 85.33\\
SeCom ~\citeyearpar{secom} & 47.99 & 21.97 & 4.12 & 21.06 & 11.68 & 21.85 & 85.47 \\
MemGAS~\citeyearpar{memgas} & 44.46 & 21.52 & 4.56 & 22.47 & 11.29 & 21.37 & 85.44\\
\method & \textbf{53.58} & \textbf{25.12} & \textbf{5.59} & \textbf{25.97} & \textbf{14.01} & \textbf{24.84} & \textbf{85.82} \\


\midrule
\multicolumn{8}{c}{\cellcolor{cyan!5}\textit{ \textbf{Gemini-2.5-flash}}} \\
\midrule
Contriever ~\citeyearpar{contriever}& 33.23 & 17.65 & 3.35 & 18.06 & 8.86 & 17.03 & 84.82 \\
SeCom ~\citeyearpar{secom} & 35.55 & 21.97 & 4.12 & 21.06 & 11.68 & 21.85 & 85.47 \\
MemGAS~\citeyearpar{memgas}& 38.37 &17.84&3.33&18.36&8.92&17.42&85.00 \\

\method &\textbf{43.47}&\textbf{23.21}&\textbf{4.81}&\textbf{23.59}&\textbf{12.44}&\textbf{22.43}&\textbf{85.81} \\

\bottomrule
\end{tabular*}
\label{tab:diff_gen}
\end{table*}

\begin{table*}[t!]
\setlength\tabcolsep{0pt}  
\small
\centering
\caption{Retrieval Performance Comparison of Different Retrievers on LoCoMo with \texttt{gpt-4o-mini-2024-07-18}.}
\begin{tabular*}{\linewidth}{@{\extracolsep{\fill}} l|ccccccc }
\toprule
\textbf{Model} &\textbf{Recall@3 }&\textbf{NDCG@3}&\textbf{Recall@5} &\textbf{NDCG@5}&\textbf{Recall@10} &\textbf{NDCG@10} \\

\midrule
\multicolumn{7}{c}{\cellcolor{cyan!5}\textit{ \textbf{Contriever}}} \\
\midrule
Contriever ~\citeyearpar{contriever}&49.95&52.15&58.31&56.29&71.80&60.91\\
MPC ~\citeyearpar{MPC} & 53.37 & 55.34 & 57.75 & 57.33 & 66.52 & 60.28 \\
RecurSum ~\citeyearpar{RecurSum} &47.23&48.99&59.01&54.58&74.97&60.07\\
SeCom ~\citeyearpar{secom} & {52.97}& {54.03}& {64.05}& {59.37}& {78.25}& {64.28}\\
MemGAS~\citeyearpar{memgas}  &\underline{56.85}&  \underline{59.22}&  \underline{67.98}& \underline{64.25}&   \underline{81.22}& \underline{68.71}  \\
\method & \textbf{69.39}  & \textbf{71.19}  & \textbf{76.33} &  \textbf{74.58} &  \textbf{86.46}  & \textbf{77.99} \\

\midrule
\multicolumn{7}{c}{\cellcolor{cyan!5}\textit{ \textbf{MPNet}}} \\
\midrule
MPNet ~\citeyearpar{song2020mpnet}&45.92&47.71&53.98&51.79&68.58&56.88\\
MPC ~\citeyearpar{MPC} &45.47&47.35&54.08&51.68&68.28&56.59\\
RecurSum ~\citeyearpar{RecurSum} & {49.50}& {51.15}& {59.47}& {56.16}& {76.64}& {61.99}\\
SeCom ~\citeyearpar{secom}
& 41.84 &  42.98 &  53.93  & 48.80  &  72.21   & 55.02 \\

MemGAS~\citeyearpar{memgas}& \underline{53.47}& \underline{54.96}& \underline{63.49}& \underline{59.95}& \underline{79.25}& \underline{65.11}\\
 
\method  & \textbf{62.74} &  \textbf{64.31} &  \textbf{70.80}  &  \textbf{68.31}  & \textbf{82.48} &  \textbf{72.21} \\

\midrule
\multicolumn{7}{c}{\cellcolor{cyan!5}\textit{ \textbf{MiniLM}}} \\

\midrule
MiniLM ~\citeyearpar{song2020mpnet}& 34.79 & 35.60 & 41.84 & 39.15 & 56.55 & 44.59\\

MPC ~\citeyearpar{MPC} &42.30&43.59&51.21&48.08&68.03&54.03\\
RecurSum ~\citeyearpar{RecurSum} &44.76&46.82&54.73&51.64& {72.16}& {57.52}\\

SeCom ~\citeyearpar{secom} 
& 39.98  & 40.99  & 51.41 &  46.64 &  70.19  & 53.31 \\

MemGAS~\citeyearpar{memgas}& \underline{46.63}& \underline{47.86}& \underline{57.25}& \underline{53.14}& \underline{72.36}& \underline{58.30}\\

\method  & \textbf{57.45} &  \textbf{59.16} &  \textbf{67.17} &  \textbf{63.67} &  \textbf{80.61} &  \textbf{68.19} \\ 
\bottomrule
\end{tabular*}

\label{tab:diff_retriever}
\end{table*}

\begin{table*}[t!]
\setlength\tabcolsep{0pt}  
\small
\caption{QA Performance Comparison of Different Retrievers on LoCoMo with \texttt{gpt-4o-mini-2024-07-18}.}
\centering
\begin{tabular*}{\linewidth}{@{\extracolsep{\fill}} l|ccccccc }
\toprule
\textbf{Model} &\textbf{4o-J}  &\textbf{F1} & \textbf{BLEU4}& \textbf{ROUGE1}&\textbf{ROUGE2}&\textbf{ROUGEL} & \textbf{BERTScore} \\

\midrule
\multicolumn{8}{c}{\cellcolor{cyan!5}\textit{ \textbf{Contriever}}} \\
\midrule
Contriever ~\citeyearpar{contriever}&40.33&15.76&2.77&16.08&7.75&15.10&84.70 \\
SeCom ~\citeyearpar{secom} & 43.45&15.28& 3.12 & 17.16 & 8.52 & 16.01 & 84.84 \\

MemGAS~\citeyearpar{memgas}  &41.07 & 17.66 &  3.61 & 18.00& 8.93 & 16.99 & 85.13 \\
\method & \textbf{48.84} & \textbf{22.66} & \textbf{5.33} & \textbf{22.92} & \textbf{12.04} & \textbf{21.68} & \textbf{85.94} \\

\midrule
\multicolumn{8}{c}{\cellcolor{cyan!5}\textit{ \textbf{MPNet}}} \\
\midrule
MPNet ~\citeyearpar{song2020mpnet} & 38.07 & 14.52 & 2.36 & 14.97 & 6.82 & 13.84 & 84.46 \\

SeCom ~\citeyearpar{secom} & 37.30 & 14.83 & 2.83 & 15.36 & 7.33 & 14.27 & 84.54 \\

MemGAS~\citeyearpar{memgas}& 37.71 & 17.43 & 3.61 & 17.83 & 8.84 & 16.81 & 85.09 \\

\method & \textbf{45.67} & \textbf{21.00} & \textbf{4.86} & \textbf{21.29} & \textbf{11.49} & \textbf{20.12} & \textbf{85.61} \\

\midrule
\multicolumn{8}{c}{\cellcolor{cyan!5}\textit{ \textbf{MiniLM}}} \\
\midrule
MiniLM~\citeyearpar{wang2020minilm} & 29.05 & 12.76 & 1.75 & 11.72 & 5.34 & 12.02 & 84.05 \\

SeCom ~\citeyearpar{secom} & 32.20 & 13.78 & 2.35 & 14.27 & 6.58 & 13.28 & 84.29 \\

MemGAS~\citeyearpar{memgas}& 32.38 & 15.92 & 3.09 & 16.47 & 7.82 & 15.41 & 85.02 \\

\method & \textbf{43.45} & \textbf{20.09} & \textbf{4.68} & \textbf{20.39} & \textbf{10.62} & \textbf{19.24} & \textbf{85.47} \\

\bottomrule
\end{tabular*}
\label{tab:diff_gen_retriever}
\end{table*}

\subsection{Ablation and Mechanism Analysis}~\label{appendix_more_exp_ablation}
\begin{table*}[t!]
\setlength\tabcolsep{0pt}  
\small
\caption{Ablation Study. Contriever is the retrieval backbone, with \texttt{gpt-4o-mini-2024-07-18} as the generator. $R$ and $G$ represents our retrieval and generation mechanism, respectively.}
\centering
\begin{tabular*}{\linewidth}{@{\extracolsep{\fill}} l|ccccccc|c }
\toprule
\textbf{Model} &\textbf{4o-J}  &\textbf{F1} & \textbf{BLEU4}& \textbf{ROUGE1}&\textbf{ROUGE2}&\textbf{ROUGEL} & \textbf{BERTScore} & \textbf{\makecell{Avg.\\Tokens}}\\

\midrule
\multicolumn{9}{c}{\cellcolor{cyan!5}\textit{ \textbf{LoCoMo}}} \\
\midrule
Baseline &40.33&15.76&2.77&16.08&7.75&15.10&84.70 & 2,348\\

+ $R$ & 43.42 & 19.62 & 3.85 & 19.86 & 10.43 & 18.68 & 85.33 & 2,685 \\

+ $R$ + $G$ & \textbf{48.84} & \textbf{22.66} & \textbf{5.33} & \textbf{22.92} & \textbf{12.04} & \textbf{21.68} & \textbf{85.94} & 1,403\\

\midrule
\multicolumn{9}{c}{\cellcolor{cyan!5}\textit{ \textbf{LongMTBench+}}} \\
\midrule

Baseline&63.54& 31.82 &7.51&33.42&18.91 & 26.53 & 87.01 & 12,045 \\

+ $R$ & 62.15 & 36.77 & 11.31 & 38.77 & 21.69 & 29.70 & 87.95 & 12,105 \\

+ $R$ + $G$ & \textbf{64.15} & \textbf{42.40} & \textbf{16.83} & \textbf{44.55} & \textbf{25.45} & \textbf{35.57} & \textbf{89.34} & 6,174\\

\midrule
\multicolumn{9}{c}{\cellcolor{cyan!5}\textit{ \textbf{LongMemEval-s}}} \\
\midrule

Baseline& 55.40 & 13.37 & 2.15 & 13.90 & 6.72 & 12.43 & 83.64 & 8,471 \\

+ $R$ & 56.00 & 14.29 & 2.20 & 14.83 & 7.21 & 13.45 & 83.85 & 8,439 \\

+ $R$ + $G$ & \textbf{57.20} & \textbf{21.06} & \textbf{4.89} & \textbf{21.80} & \textbf{10.77} & \textbf{20.16} & \textbf{85.54} & 4,302\\

\midrule
\multicolumn{9}{c}{\cellcolor{cyan!5}\textit{ \textbf{LongMemEval-m}}} \\
\midrule
Baseline &42.80&11.88&1.66&12.63&5.51&11.11&83.31 & 8,274\\

+ $R$ & 46.00 & 12.86 & 1.92 & 13.54 & 6.30 & 12.15 & 83.60 & 8,357 \\

+ $R$ + $G$  & \textbf{46.60} & \textbf{18.09} & \textbf{3.96} & \textbf{18.88} & \textbf{9.03} & \textbf{17.40} & \textbf{85.10} & 4,261 \\ 

\bottomrule
\end{tabular*}
\label{tab:appendix_ablation}
\end{table*}

\begin{figure*}[h]
    \centering
    \includegraphics[width=1.0\textwidth]{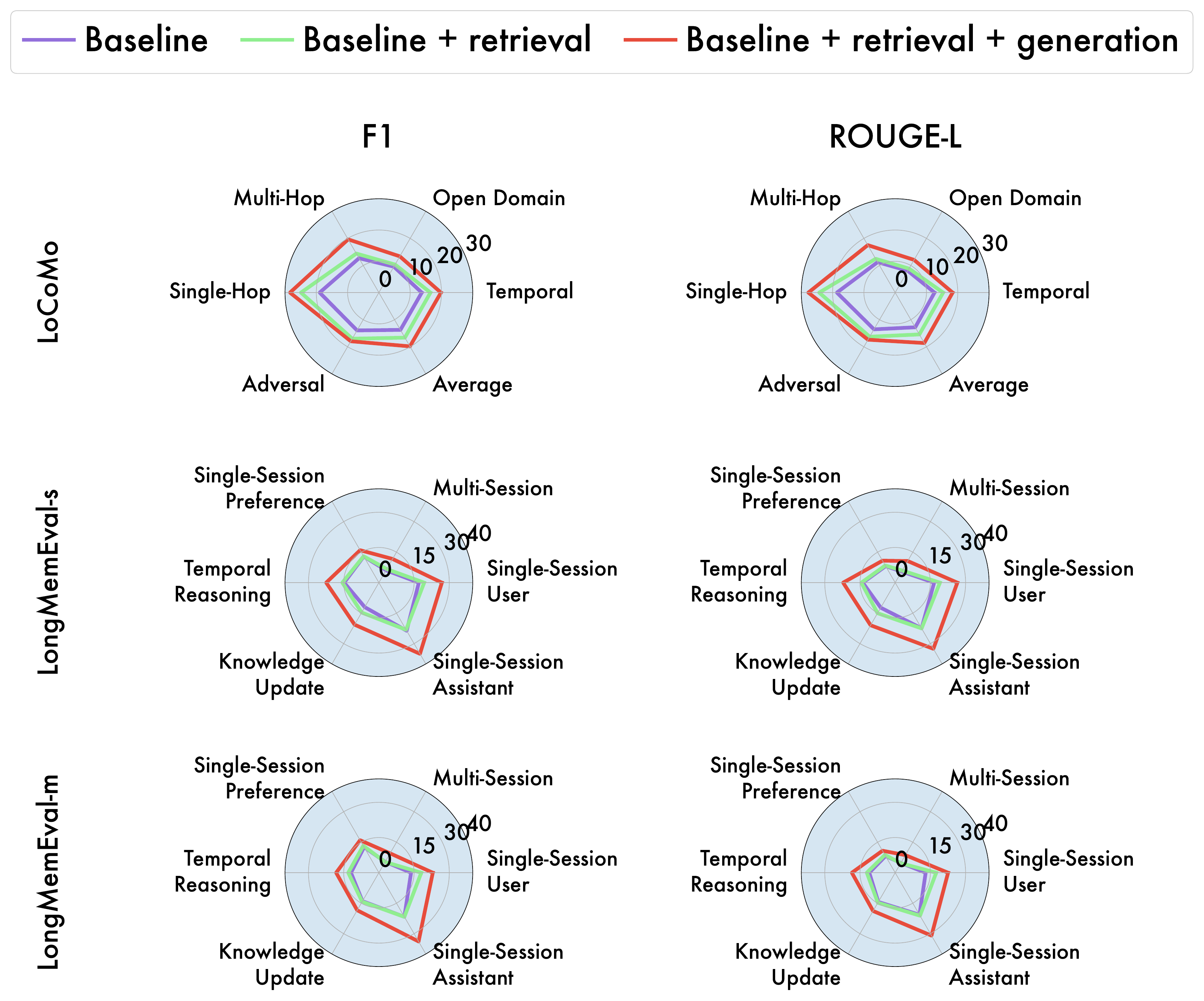}
    \caption{Ablation study on different query settings using \textit{gpt-4o-mini-2024-07-18} and Contriever.}
    \label{fig:ablation_appendix}
\end{figure*}

We provide complete ablation results on four long-term conversational benchmarks, as presented in Tab.~\ref{tab:appendix_ablation} and the different query ablation in Fig~\ref{fig:ablation_appendix}. 
We observe similar analysis as in the main text. 


\clearpage
\section{Prompts}~\label{appendix_prompts}

The following query-aware pruning prompt selectively retains content relevant to the input query while preserving original token fidelity.

\begin{tcolorbox}[colback=gray!10,colframe=black,title=Question Answer]
You are an intelligent dialog bot. You will be shown a "Fused Historical Event" which contains all the consolidated details relevant to your question. Filter the information to extract only the parts directly relevant to the Question. Preserve original tokens, do not paraphrase. Remove irrelevant turns, redundant info, and non-essential details.

\textbf{Fused Historical Event:} <fused\_event> \\
\textbf{Question Date:} <question date> \\
\textbf{Question:} <question> \\
\textbf{Answer:}
\end{tcolorbox}

For QA task, we follow~\citep{memgas} and prompt the model with the following template to produce a concise, history-based response.

\begin{tcolorbox}[colback=gray!10,colframe=black,title=Question Answer]
You are an intelligent dialog bot. You will be shown History Dialogs. Please read, memorize, and understand the given Dialogs, then generate one concise, coherent and helpful response for the Question.

\textbf{History Dialogs:} <retrieved texts>\\
\textbf{Question Date:} <question date>\\
\textbf{Question:} <question>

\end{tcolorbox}

For evaluation, we prompt GPT-4o with the template from~\citep{zheng2023judging} to judge whether the model’s output matches the correct answer, as detailed below.

\begin{tcolorbox}[colback=gray!10,colframe=black,title=GPT-4o-as-Judge]
I will give you a question, a reference answer, and a response from a model. Please answer <yes> if the response contains the reference answer. Otherwise, answer <no>. If the response is equivalent to the correct answer or contains all the intermediate steps to get the reference answer, you should also answer <yes>. If the response only contains a subset of the information required by the answer, answer <no>.

\textbf{User Question:} <question>\\
\textbf{Correct Answer:} <answer>\\
\textbf{Model Response:} <response>\\
Is the model response correct? Answer <yes> or <no> only.

\end{tcolorbox}
\section{More Discussion}

\subsection{Current Limitation}
While our minimalist memory framework simplifies the retrieval process by effectively leveraging raw conversational history, it remains subject to the inherent limitations of non-parametric memory systems, specifically regarding privacy, interpretability, and security risks. Long-term agentic memory inevitably stores persistent and potentially sensitive user-specific content, including interaction histories and behavioral traces. 
Although retrieving such information is essential for generating personalized, high-quality responses, it inherently introduces risks of data leakage and excessive information retention. 
Furthermore, current systems are constrained by limited interpretability, lacking robust diagnostic tools to precisely trace which specific memory entries were retrieved and how they directly influenced the large language model’s generative process.

\subsection{Future Direction}
Future research will explore two directions. First, to transition from a passive retrieval framework to a proactive memory system, memory management should be integrated into the LLM's generation process via outcome-driven Reinforcement Learning~\citep{yue2026mem,lei2025federated,lei2025game}. Fine-tuning the agent~\cite{yu2025co, wu2025mas4poi} to autonomously emit structured memory operations (e.g., add, update, delete) during inference. Second, the integration of privacy-preserving mechanisms such as differential privacy or selective forgetting is essential to ensure data security. These advancements will be pivotal for maintaining user trust while enhancing the long-term utility of memory systems.

\section{LLM Disclosure}

Following the conference guidelines regarding Generative AI models (GenAI, include large language model, a.k.a. LLM), we disclose that GenAI was utilized solely to improve sentence clarity and grammatical correctness. All aspects of conceptual development, experimental methodology, data analysis, and core manuscript content were independently produced by the authors without GenAI assistance.

\clearpage
\newpage

\end{document}